\documentclass[sigconf,screen]{acmart}

\usepackage{times}
\usepackage{epsfig}
\usepackage{graphicx}
\usepackage{amsmath}
\usepackage{booktabs}
\usepackage{tikz}
\usepackage{comment}
\usepackage{color}
\usepackage{subfigure}
\usepackage{multirow}
\usepackage{diagbox}
\usepackage{rotating}
\usepackage{booktabs}
\usepackage{colortbl}
\usepackage{overpic}
\usepackage{makecell}
\usepackage{textcomp}
\usepackage{contour}
\usepackage{algorithm}
\usepackage{algorithmic}

\AtBeginDocument{%
  }

\setcopyright{acmcopyright}
\copyrightyear{2023}
\acmYear{2023}
\acmDOI{XXXXXXX.XXXXXXX}

\acmConference[MM'23]{Proceedings of the 31st ACM International Conference on Multimedia (MM '23}{October 2023}{Ottawa, Canada}
\acmBooktitle{Proceedings of the 31st ACM International Conference on Multimedia (MM '23),October 29 -- November 03,2023, Ottawa, Canada}

\acmPrice{15.00}
\acmISBN{978-1-4503-XXXX-X/18/06}

\acmSubmissionID{506}

\newcommand{\sm}[1]{{\textcolor{magenta}{#1}}}

\newcommand{\figref}[1]{Fig. \ref{#1}}
\newcommand{\tabref}[1]{Table \ref{#1}}
\newcommand{\eqnref}[1]{Eq. (\ref{#1})}
\newcommand{\secref}[1]{Sec. \ref{#1}}

\newcommand{\ourmodel}{UDUN}

\renewcommand{\tabcolsep}{.5mm}

\newcommand{\tabincell}[2]{
\begin{tabular}{@{}#1@{}}#2\end{tabular}
}

\definecolor{mygray}{gray}{.95}
\definecolor{myRed}{RGB}{219, 68, 55}
\definecolor{myGreen}{RGB}{15, 157, 88}
\definecolor{myBlue}{RGB}{66, 133, 244}
\newcommand{\tr}[1]{{\textcolor{myRed}{\textbf{#1}}}}
\newcommand{\tg}[1]{{\textcolor{myGreen}{\textbf{#1}}}}
\newcommand{\tb}[1]{{\textcolor{myBlue}{\textbf{#1}}}}

\newcommand{\eg}{\emph{e.g.}}
\newcommand{\ie}{\emph{i.e.}}
\newcommand{\etal}{\emph{et al.}}

\graphicspath{{./Imgs/}}

\renewcommand\footnotetextcopyrightpermission[1]{}
\begin{document}

\title{Unite-Divide-Unite: Joint Boosting Trunk and Structure for High-accuracy Dichotomous Image Segmentation}


\author{Jialun Pei}
\authornote{Both authors contributed equally to this research.}
%\orcid{0000-0002-2630-2838}
 \affiliation{School of Computer Science and Engineering, 
   \institution{The Chinese University of Hong Kong}
% \streetaddress{P.O. Box 1212}
  \city{Shatin}
  \state{NT}
  \country{Hong Kong}
}
\email{jialunpei@cuhk.edu.hk}

\author{Zhangjun Zhou}
\authornotemark[1]
 \affiliation{School of Software Engineering, 
   \institution{Huazhong University of Science and Technology}
\streetaddress{Wuhan, China}
  \city{Wuhan}
%  \state{NT}
  \country{China}
}
\email{zhouzhangjun@hust.edu.cn}

\author{Yueming Jin}
 \affiliation{School of Electrical and Computer Engineering, 
   \institution{National University of Singapore}
% \streetaddress{P.O. Box 1212}
  \city{}
%  \state{NT}
  \country{Singapore}
}
\email{ymjin@nus.edu.sg}

\author{He Tang}
\authornote{Corresponding author: He Tang (E-mail: hetang@hust.edu.cn)}
 \affiliation{School of Software Engineering, 
   \institution{Huazhong University of Science and Technology}
% \streetaddress{P.O. Box 1212}
  \city{Wuhan}
%  \state{NT}
  \country{China}
}
\email{hetang@hust.edu.cn}

\author{Pheng-Ann Heng}
%\orcid{0000-0002-2630-2838}
 \affiliation{School of Computer Science and Engineering, 
   \institution{The Chinese University of Hong Kong}
% \streetaddress{P.O. Box 1212}
  \city{Shatin}
  \state{NT}
  \country{Hong Kong}
}
\email{pheng@cse.cuhk.edu.hk}


% % % %图1得展示内部区域结构识别清晰的例子（如椅子）对比实验的定性比较也是（pipeline的摩天轮好像看不出内部结构预测的困难，但是展示了trunk和structure的解耦学习）。
% \begin{figure*}[p!]
%     \centering
%     \includegraphics[width=\linewidth]{Imgs/Fig1.pdf}
%     \caption{Motivation}\label{fig:motivation}
% \end{figure*}

\begin{abstract}

High-accuracy Dichotomous Image Segmentation (DIS) aims to pinpoint category-agnostic foreground objects from natural scenes. 
%Unlike high-resolution semantic-specific binary image segmentation, DIS involves identifying the accurate foreground area while rendering detailed object structure.
The main challenge for DIS involves identifying the highly accurate dominant area while rendering detailed object structure. 
However, directly using a general encoder-decoder architecture may result in an oversupply of high-level features and neglect the shallow spatial information necessary for partitioning meticulous structures. 
To fill this gap, we introduce a novel \textbf{Unite-Divide-Unite Network (\ourmodel)} that restructures and bipartitely arranges complementary features to simultaneously boost the effectiveness of trunk and structure identification.
The proposed~\ourmodel~proceeds from several strengths.
First, a dual-size input feeds into the shared backbone to produce more holistic and detailed features while keeping the model lightweight.
Second, a simple \emph{Divide-and-Conquer Module (DCM)} is proposed to decouple multiscale low- and high-level features into our structure decoder and trunk decoder to obtain structure and trunk information respectively.
Moreover, we design a \emph{Trunk-Structure Aggregation module (TSA)} in our union decoder that performs cascade integration for uniform high-accuracy segmentation.
As a result, \ourmodel~performs favorably against state-of-the-art competitors in all six evaluation metrics on overall DIS-TE, \ie, achieving 0.772 weighted F-measure and 977 HCE. Using 1024$\times$1024 input, our model enables real-time inference at 65.3 \emph{fps} with ResNet-18. 
% The source code will be made available.
The source code is available at \sm{https://github.com/PJLallen/UDUN.}

\end{abstract}

%   <concept>
%       <concept_id>10010147</concept_id>
%       <concept_desc>Computing methodologies</concept_desc>
%       <concept_significance>300</concept_significance>
%       </concept>
%   <concept>
%       <concept_id>10010147</concept_id>
%       <concept_desc>Computing methodologies</concept_desc>
%       <concept_significance>500</concept_significance>
%       </concept>
%   <concept>
%       <concept_id>10010147.10010371.10010382.10010383</concept_id>
%       <concept_desc>Computing methodologies~Image processing</concept_desc>
%       <concept_significance>500</concept_significance>
%       </concept>
%  </ccs2012>

% \ccsdesc[300]{Computing methodologies}
% \ccsdesc[500]{Computing methodologies}
% \ccsdesc[500]{Computing methodologies~Image processing}


\begin{teaserfigure}
  \includegraphics[width=\textwidth]{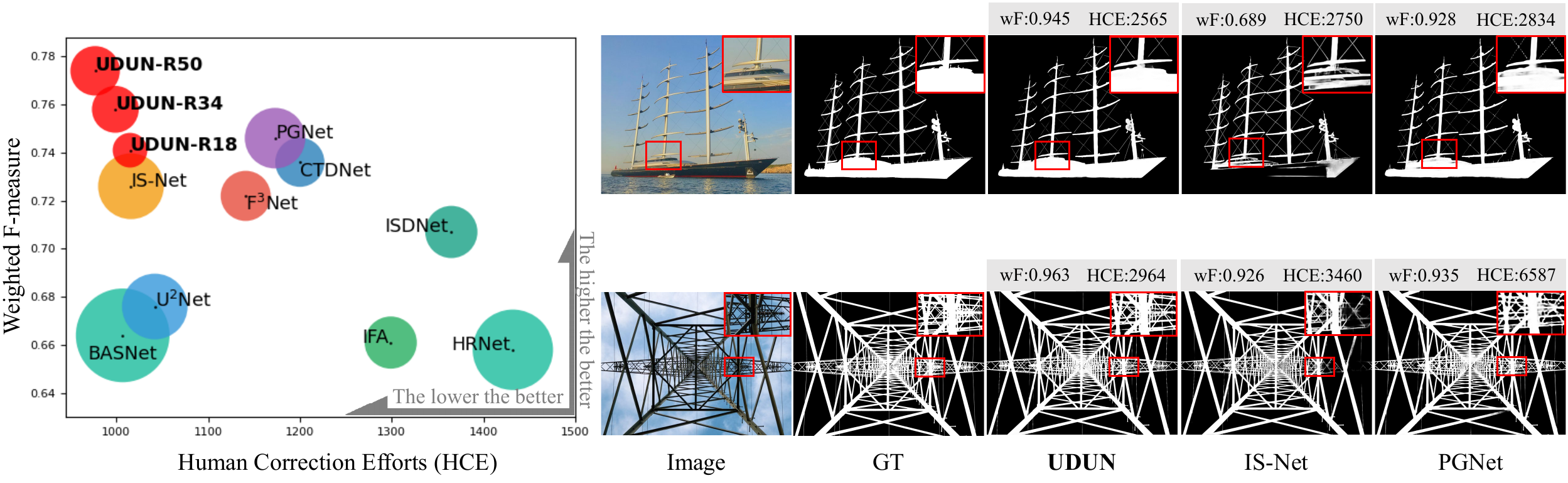}
  \put(-405,-10){{(a)}}
  \put(-160,-10){{(b)}}
  \caption{Dichotomous image segmentation (DIS) encourages the methods to boost both weighted F-measure and HCE. (a) Comparisons of the proposed~\ourmodel\ with competitors in terms of weighted F-measure, HCE and number of parameters on the DIS-TE test set. The area of each circle represents the number of parameters of the respective model. UDUN achieves superior performance in all metrics. (b) Visual comparison of corresponding metrics. UDUN can take into account both the detailed structure and integrity of targets.}
  \Description{E}
  \label{fig:motivation}
\end{teaserfigure}

% Dichotomous image segmentation (DIS) encourages the methods to boost both weighted F-measure and HCE. (a) Comparisons of the proposed~\ourmodel\ with other competitors in terms of weighted F-measure, human correction efforts (HCE) and number of parameters on the DIS-TE test set. The area of each circle corresponds to the number of parameters of the respective model. Our method achieves superior performance in all metrics. (b) Visual comparison of corresponding metrics. UDUN can take into account both the detailed structure and integrity of targets.

\maketitle

\section{Introduction}
%With the significant improvement in the representational capacity of deep neural networks, modern semantic segmentation \cite{} and class-agnostic segmentation \cite{} methods have obtained high accuracy at predict the general mask of the foreground. However, high-demand scenarios like intelligent healthcare and intelligent construction expect the segmentation results with both trunk and structure preserving. Recently, Qin $\etal$ \cite{qin2022highly} introduced a new category-agnostic dichotomous image segmentation (DIS) task to produce highly accurate mask of foreground area. Moreover, they proposed a new human correction efforts (HCE) metric to evaluate structural accuracy of segmentation other than traditional trunk accuracy metrics like maximal F-measure \cite{achanta2009frequency}, mean absolute error \cite{perazzi2012saliency}, mean enhanced alignment measure \cite{fan2018enhanced}, etc. The study of DIS is promising for extensive applications like medical image analysis \cite{}, AR/VR \cite{}, etc.
With the prominent developments in the representational capacity of deep neural networks, modern semantic segmentation~\cite{shen2022high,guo2022isdnet} and class-agnostic segmentation methods~\cite{tang2021disentangled,pang2022zoom} have yielded high accuracy in covering desired areas of the foreground. 
However, high-demand scenarios like intelligent healthcare and smart construction expect segmentation to preserve both the integrity of trunk and detailed structure~\cite{zhuge2022salient}. As two sides of a coin, the trunk involves in the dominant area and the structure focuses on fine-grained internal and external edges of objects.
Recently, Qin \etal~\cite{qin2022highly} introduced a new class-agnostic dichotomous image segmentation (DIS) task to produce a highly accurate mask of foreground objects.
Moreover, they proposed a new human correction efforts (HCE) metric that measures the difference between predictions and realistic applications by estimating the need for human interventions to calibrate interior structures and exterior boundaries.
It indicates that the HCE metric is more sensitive to the structural refinement of the segmentation map other than traditional accuracy metrics like weighted F-measure~\cite{achanta2009frequency}, mean absolute error~\cite{perazzi2012saliency}, and mean enhanced alignment measure~\cite{fan2018enhanced}. 
Therefore, compared to general task-specific object segmentation~\cite{fan2022salient,fan2021concealed}, DIS has a greater challenge since it demands simultaneous attention to the integrity of the target and the finer structure.
Research into DIS is promising for extensive applications such as robotic intelligence~\cite{zhao2022trasetr}, medical image analysis~\cite{valanarasu2022unext}, and AR/VR applications~\cite{tian2022kine}.

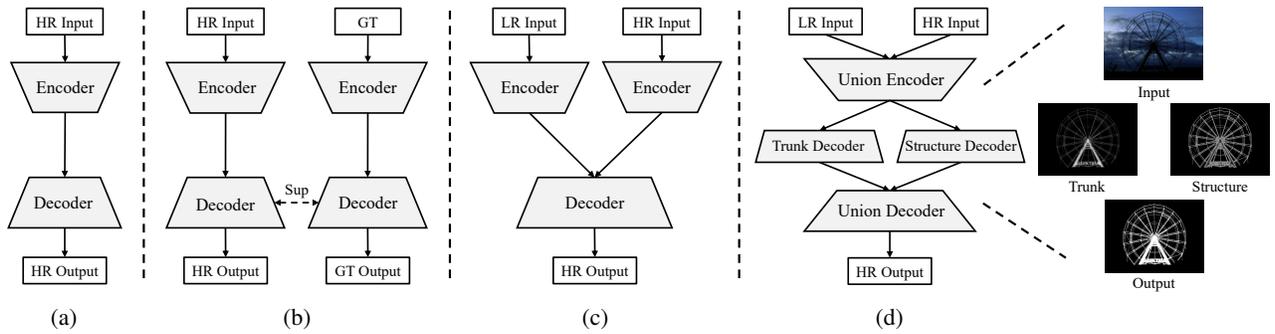
\begin{figure*}[t!]
\centering
\includegraphics[width=0.95\linewidth]{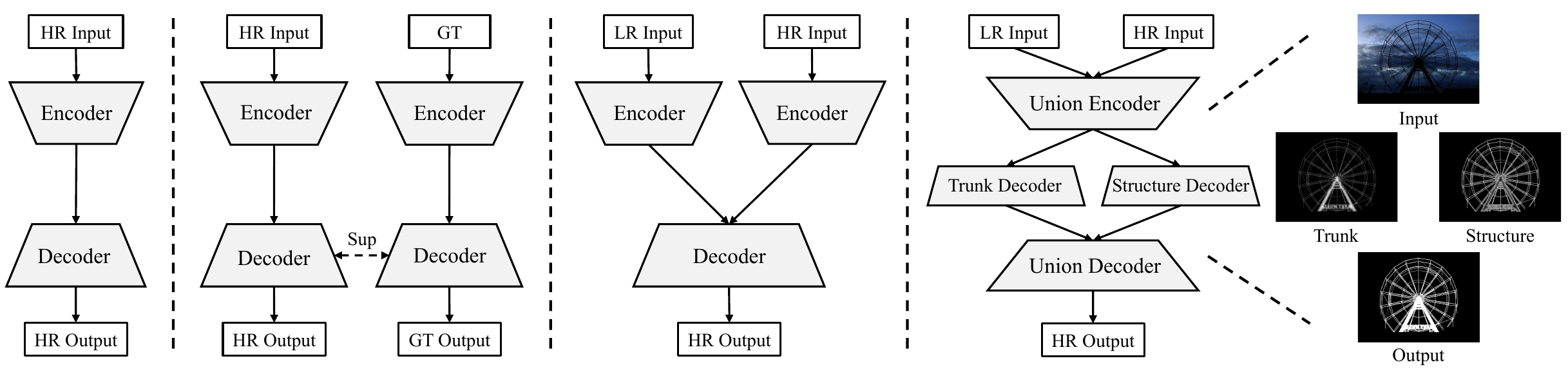}
  \put(-463,-10){{(a)}}
  \put(-375,-10){{(b)}}
  \put(-262,-10){{(c)}}
  \put(-151,-10){{(d)}}
\caption{Architecture comparisons of high-accuracy image segmentation models. (a) General FCN-based encoder-decoder segmentation model; (b) Two-stage intermediate supervision model for DIS~\cite{qin2022highly}; (c) Dual-size input two encoder model~\cite{xie2022pyramid}; (d) The proposed dual-size input unite-divide-unite model for DIS. HR and LR represent high and low resolution, respectively.}\label{fig:Cmp}
\end{figure*}

% %图1得展示内部区域结构识别清晰的例子（如椅子）对比实验的定性比较也是（pipeline的摩天轮好像看不出内部结构预测的困难，但是展示了trunk和structure的解耦学习）。
% \begin{figure*}[t!]
% \centering
% \includegraphics[width=\linewidth]{Imgs/Fig1.pdf}
% \caption{Motivation}\label{fig:motivation}
% \end{figure*}

%Various works \cite{} have been proposed to promote high-accuracy image segmentation with respect to the encoder-decoder framework.
%(Fig. \ref{fig:Cmp}(a)). IS-Net (Fig. \ref{fig:Cmp}(b)) \cite{qin2022highly} trains the network in a two-stage manner with high-dimensional intermediate supervision. PGNet (Fig. \ref{fig:Cmp}(c)) \cite{xie2022pyramid} adopts dual-input and dual-encoder to extract features from different resolution images, then interacts the features with a cross-model grafting module. 
%Though some recently proposed methods, like IS-Net \cite{qin2022highly} and PGNet \cite{xie2022pyramid} have achieved good performance for DIS task, they still suffer from two major drawbacks as shown in Fig. \ref{fig:motivation}: 1) the segmentations with complex internal structures are inaccurate; 2) the weighted F-measure and HCE of non-convex objects can hardly be simultaneously boosted. This can be attributed to: 1) the features of later stages are high-level and lack of details for structure segmentation; 2) segmenting trunk and structure in high-accuracy are conflict in some extent.
Various prominent works~\cite{cheng2020cascadepsp,zhang2021looking} have been spawned to facilitate high-accuracy image segmentation with respect to the encoder-decoder framework. 
IS-Net~\cite{qin2022highly} provides the first solution to the DIS task by cascading multiple U-Net structures~\cite{ronneberger2015u} with intermediate supervision.
PGNet~\cite{xie2022pyramid} adopts a dual-branch architecture and embraces a cross-model grafting module to detect high-resolution salient objects.
Despite the satisfactory performance achieved by previous works, two major issues remain unresolved for high-accuracy DIS:
(1) as demonstrated in~\figref{fig:motivation}(a), it is difficult to simultaneously improve the quality of the holistic segmentation of the target (refer to weighted F-measure) and detailed structural segmentation of non-convex objects (refer to HCE); (2) as shown in~\figref{fig:motivation}(b), high-resolution targets with intricate structures are challenging to segment properly and entirely.
The underlying reasons may be: (i) existing methods often struggle to simultaneously reconcile the dominant area and internal details of the object for high-accuracy segmentation; (ii) features extracted from deeper layers of the FCN-based encoder have a higher level of aggregation, resulting in a lack of low-level structural and spatial information.

%To address the above-mentioned issues, we propose a dual-size input unite-divide-unite network (\ourmodel) to disentangle the trunk and structure segmentation for high-accuracy dichotomous image segmentation. Union encoder is proposed to extract multi-level features with shared backbone for parameter-saving. Thereinto, Divide-and-conquer Module (DCM) splits multi-level features from the dual-size input, then recombine the features to prevent high-level features for structure segmentation. The trunk decoder and structure decoder are designed to segment core mask and interior structures of objects respectively, this design alleviates the conflict between trunk and structure segmentations. Finally, Union Decoder is proposed to unite both structure and trunk prediction for the final dichotomous image segmentation.
To overcome the above-mentioned issues, we propose a dual-size input unite-divide-unite network (\ourmodel) to disentangle the trunk and structure segmentation for high-accuracy DIS. 
Concretely, we introduce a union encoder to extract multi-layer features by feeding larger and smaller inputs to a shared backbone for parameter-saving. 
Thereinto, the divide-and-conquer module (DCM) recombines two groups of multi-level features for efficient divisional optimization. 
The trunk decoder and structure decoder are designed to separately refine and fuse the dominant area and detailed structure information. Finally, a union decoder is proposed to integrate and unify the structure and trunk information for the final dichotomous segmentation. 
The unite-divide-unite strategy effectively alleviates the conflict between trunk and structure segmentation. 
% As shown in Fig. \ref{fig:Cmp}, the network of high-accuracy image segmentation can be build upon a general FCN-based encoder-decoder model (Fig. \ref{fig:Cmp}(a)). Compare with IS-Net \cite{qin2022highly} (Fig. \ref{fig:Cmp}(b)), the proposed \ourmodel (Fig. \ref{fig:Cmp}(d)) requires only one-stage training and not heavily rely on intermediate supervision, which makes \ourmodel more easier training. Different from PGNet \cite{xie2022pyramid} (Fig. \ref{fig:Cmp}(c)), \ourmodel (Fig. \ref{fig:Cmp}(d)) requires single united encoder and with disentangled decoders, which makes the model more parameter-saving and higher accuracy.
\figref{fig:Cmp} shows a comparison of high-accuracy image segmentation architectures built upon a general FCN-based encoder-decoder model (\figref{fig:Cmp}(a)). 
Compared to IS-Net~\cite{qin2022highly} (\figref{fig:Cmp}(b)), the proposed~\ourmodel~(\figref{fig:Cmp}(d)) framework requires only one stage for training and relies not heavily on intermediate supervision, which makes the~\ourmodel~ easier to converge.
Different from PGNet~\cite{xie2022pyramid} (\figref{fig:Cmp}(c)), \ourmodel~ employs a single union encoder with disentangled decoders, which contributes to the efficiency and higher accuracy of our model. 
Extensive experiments demonstrate that~\ourmodel~achieves state-of-the-art performance on the DIS-TE test set~\cite{qin2022highly}, in terms of both the integrity of the object and the fineness of the structure, \eg, 0.831, 0.892, and 977 on the $F_\beta^\text{max}$, $E_{\phi}^\text{m}$, and HCE$_{\gamma}$ respectively.

% Overall framework of the proposed~\ourmodel. \ourmodel is organized as: a union encoder with the divide-and-conquer module (DCM), a structure decoder, a trunk decoder, and a union decoder with the trunk-structure aggregation module (TSA) and the mask-structure aggregation module (MSA).

%贡献
%单阶段的DIS解决方案:解决了前人工作忽略的问题
%我们采用了UEncoder 以减少参数的方式去获取丰富的多尺度的特征
%并且，使用Divide-and-Conquer Module 的策略重组高级特征和低级特征，分别用于预测前景目标的trunk   和 structure。moreover，TSA被设计用来融合两种互补的特征，以此得到高精度的预测图。
%所有DISTE测试集上，我们的方法在六个评价指标上attains a superior performance，并且都达到了sota 和实时的速度with不同的backbone ResNet-18 ResNet-34 ResNet-50。
Our main contributions can be summarized as follows:
\begin{itemize}
    %    \item We propose a dual-size input unite-divide-unite network (\ourmodel) for high-accuracy dichotomous image segmentation. It not only produces finer segmentation for complex objects, but also alleviates the conflict between trunk and structure segmentation.
    \item We propose a novel unite-divide-unite network (\ourmodel) for high-accuracy DIS. This framework contains a union encoder to efficiently obtain rich multiscale features, a trunk and structure decoder to refine and merge multi-level trunk and structure features respectively, and a union decoder to aggregate cross-structural information. \ourmodel~provides finer segmentation of complex objects and joint optimization of trunks and structures.
    % \item UDUN disentangles the trunk and structure segmentation from DIS to boost both accuracies. In this way, we introduce a union encoder to obtain rich multiscale features in a parameter-saving manner, it contains a Divide-and-Conquer Module to recombine high- and low-level features, which are used for predicting the complementary trunk and structure of the foreground, respectively. Moreover, Trunk-Structure Aggregation module is designed to fuse these complementary features, resulting in high-accuracy segmentation maps.
    \item A divide-and-conquer module (DCM) is embedded in our union encoder to recombine high- and low-level features for capturing complementary trunk and structure cues respectively. Moreover, we design a trunk-structure aggregation module (TSA) to sufficiently integrate these complementary features for a unified mask feature.
    \item \ourmodel~achieves state-of-the-art performance in terms of all six metrics on DIS-TE at real-time speed, demonstrating consistent advancements in trunk and structure accuracy.
\end{itemize} 

\section{Related Works}

\subsection{High-resolution Semantic Segmentation}
% 说明高分辨语义分割任务的目的，总结几种相关模型的处理方式（基于全卷积网络模型，由粗到细，级联方式等等）。然而，他们都是基于语义相关的目标来进行精细化分割，但都忽略了在语义区域中结构化的细节。为此，我们提出了一个XX模型，针对结构复杂度较高的目标，可以进一步提取更加精细化的细节特征以达到更高质量的分割。

%高分辨率语义分割是为了在高分辨率的样本上识别出语义相关的目标。
%一些方法,利用了降采样或者裁剪成patch的方法得到低分辨率的特征来高效率的定位语义。另一方面。
%为了直接处理高分辨率样本，一些轻量化的模型[,,]被设计出来以便处理高分辨率特征。然而，前者为了追求语义定位而牺牲了空间细节，后者往往存在感受野不足的问题.
%BiSeNetV1  v2,ISDNet(这些都是实时的，不该提增加了参数量)
% 使用一个浅层分支来提取高分辨率输入低级特征，一个深层分支从低分辨率输入提取高级特征。这些工作采用了双输入的策略来提高准确率并达到实时的速度。
%然而，他们都是基于语义相关的目标来进行精细化分割，但都忽略了在语义区域中结构化的细节。
% 因此，我们设计了一个简洁的双输入高分辨率分割模型，它具有轻量化的结构并可以准确的同时捕捉语义特征和细节特征。

High-resolution semantic segmentation aims to identify the classes of each pixel in high-resolution images. 
Numerous approaches~\cite{zeng2019towards,chen2019collaborative,li2021contexts}  leverage downsampling or patch cropping techniques to efficiently locate semantics using low-resolution features. In order to directly handle high-resolution samples, some lightweight models~\cite{pohlen2017full,romera2017erfnet,chaurasia2017linknet}  have been designed to handle and preserve high-resolution features. However, the former sacrifices spatial details in pursuit of semantic localization, while the latter often suffers from insufficient receptive field. Recently, several methods~\cite{yu2018bisenet,fan2021rethinking,guo2022isdnet} employed a lightweight branch to extract low-level features from the high-resolution input together with a heavyweight branch to extract high-level features from the low-resolution input. 
These works adopted a dual-input strategy to improve accuracy and achieve real-time speeds. 
For example, ISDNet~\cite{guo2022isdnet} integrated deep and shallow networks to construct an efficient segmentation model for remote sensing fields.
% in remote sensing field.
%However, they are all based on semantically-predefined objectives for fine-grained segmentation, while ignoring structural details within semantic regions. Therefore, we propose a concise dual-input high-resolution segmentation model with a lightweight structure that capture high-level semantic features and low-level detail features simultaneously.

Nevertheless, such methods all rely on semantically-defined objectives for fine-grained segmentation, while ignoring structural details within semantic regions. 
Therefore, we develop a concise dual-input high-resolution segmentation model with a lightweight architecture that collaboratively optimizes high-level semantic features and low-level detail features. 

\subsection{High-quality Class-agnostic Segmentation}
High-quality class-agnostic segmentation identifies both the task-specific foreground area and fine boundaries of the object, without predicting the class of each pixel~\cite{tang2021disentangled}.
Several works~\cite{qin2019basnet,tang2021disentangled,lin2021real} decomposed the task into two sequential stages: coarse localization and fine-grained optimization.
Generally, they employed one network to obtain coarse predictions and then another network to refine the erroneous regions. 
Toward higher-quality segmentation maps, consequential refinement techniques~\cite{cheng2020cascadepsp,yu2021mask,shen2022high} have been proposed that focus directly on refining coarse predictions. For instance, CRM~\cite{shen2022high} continuously aligned the feature map with the refinement target and aggregated features to reconstruct these image details.

% it can be implemented by concatenating the original image with the coarse prediction from other networks.}

The aforementioned task focuses solely on producing high-quality segmentations of foreground boundaries, without considering the fine details and complicated structures within the object. 
Recently, Qin~\etal~\cite{qin2022highly} proposed a new task, called highly accurate Dichotomous Image Segmentation (DIS). This task requires splitting finer details and more complex internal and external structures. 
To tackle these challenges, they developed a novel solution called IS-Net, which is based on the structure of the U$^\text{2}$-Net~\cite{qin2020u2} optimized to handle the high-resolution input and adopted an intermediate supervision strategy to reduce the risk of overfitting. 
Unlike the coarse-to-fine paradigm~\cite{qin2019basnet,tang2021disentangled,lin2021real} or the two-stage training style~\cite{qin2022highly}, we introduce a one-stage DIS model that can simultaneously boost accuracy in terms of trunk and structure (refer to \figref{fig:Cmp}(d)).
%we propose a novel "unite-divide-unite" approach. Specifically, our method involves aggregating dual-resolution inputs, passing them through a shared backbone, and then separating low-level features and high-level features. The trunk features and structure features are then predicted separately using the respective feature sets, followed by a structure features purify operation that enables precise segmentation of both the region and structured details of the target simultaneously.

\begin{figure*}[t!]
\centering
\includegraphics[width=\linewidth]{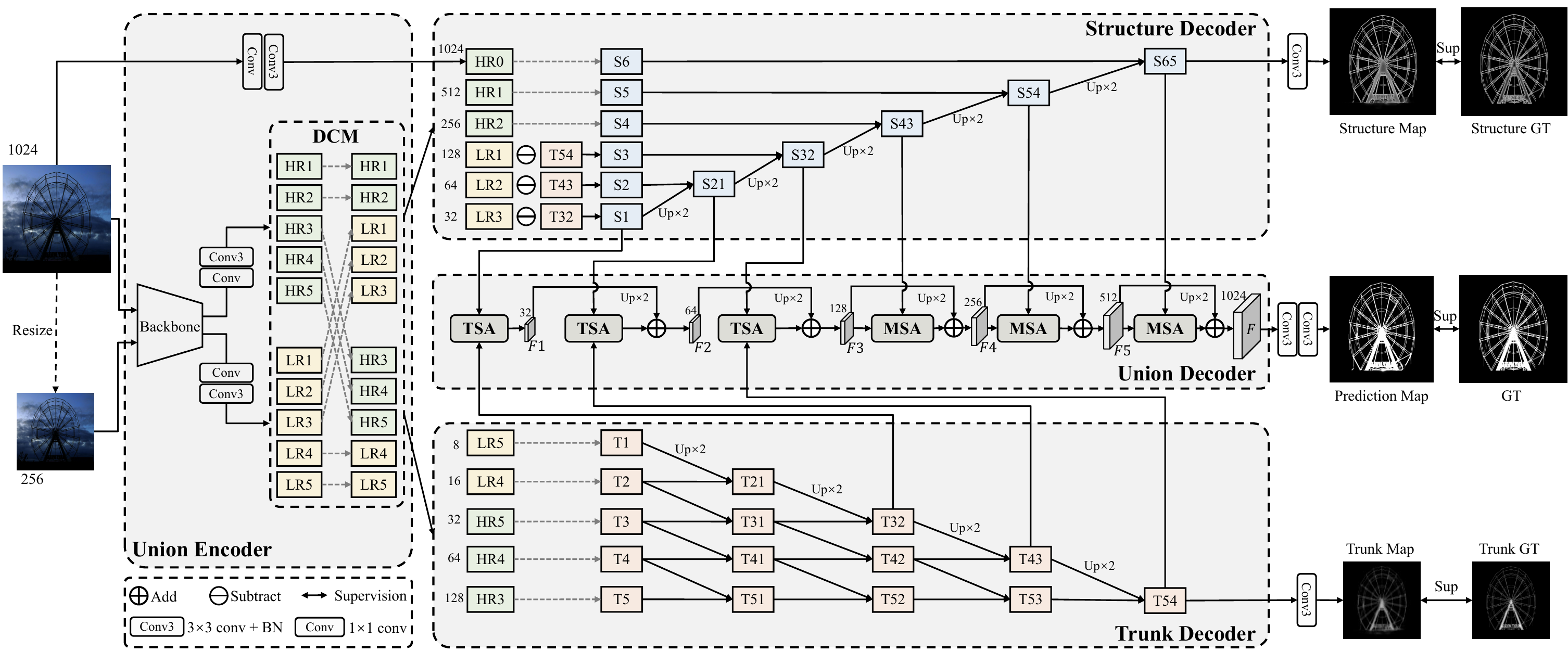}
\caption{Architecture of our~\ourmodel. \ourmodel\ includes: union encoder with divide-and-conquer module (DCM), structure decoder, trunk decoder, and union decoder with trunk-structure aggregation (TSA) and mask-structure aggregation (MSA) modules.}\label{fig:model}
\end{figure*}

\section{Proposed \ourmodel}

\subsection{Overall Architecture}

As illustrated in \figref{fig:model}, the main components of the proposed~\ourmodel\ include:
(1) A union encoder that contains a shared backbone with the dual-size input to efficiently extract multiscale low- and high-level feature representations and a divide-and-conquer module (DCM) to assemble desired features to optimize the input of different decoders. 
(2) A trunk decoder to refine trunk information through multiscale cascade upsampling.
(3) A structure decoder to capture and integrate detailed structure features.
(4) A union decoder that is designed to merge multi-level trunk and structure features to generate the binary mask. Within this, the trunk-structure aggregation module (TSA) is proposed to robustly integrate cross-structure features.

\subsection{Union Encoder}

To better extract the trunk and structure information more adequately, our union encoder adopts a dual-size input to produce multiscale features over a wider range.
In accordance with IS-Net~\cite{qin2022highly}, the larger-size input is set to 1024$\times$1024. 
The input image, denoted as $I_{h}\in\mathbb{R}^{1024\times 1024\times 3}$, is directly resized to obtain a smaller-size input $I_{l}\in\mathbb{R}^{256\times 256\times 3}$. 
To mitigate the computational cost incurred by two resolution inputs, we utilize a shared backbone (defaulted to ResNet-50~\cite{he2016deep}) to extract double group multiscale features, denoted as $\{HRi|i=1, 2, 3, 4, 5\}$ from $I_{h}$ and $\{LRi|i=1, 2, 3, 4, 5\}$ from $I_{l}$. 
It is worth noting that we apply two convolution blocks (1$\times$1 convolution, 3$\times$3 convolution, and batch normalization (BN)) to reduce the channels of $HRi$ and $LRi$ to 32 and 64 respectively for lightening the model parameters.

Considering the tailor-made treatment of structure and trunk cues, it is essential to divide and recombine two sets of features $HRi$ and $LRi$ for joint optimization.
Toward this goal, we introduce a concise divide-and-conquer module (DCM) to reorganize two groups of features. 
As shown in the DCM part of \figref{fig:model}, the main principles of division in DCM include two aspects. 
First, higher-level features with smaller scales are gathered to process trunk information, as they contain more semantic knowledge that is desirable for locating the trunk region of the object. 
Second, identifying object structures requires lower-level features with high resolution to provide more detailed texture information.
Therefore, we redivide features from the backbone into two groups: $\{HR3, HR4, HR5, LR4, LR5\}$ for the trunk decoder and $\{HR1, HR2, LR1, LR2, LR3\}$ for the structure decoder.
Additionally, to supplement larger resolution low-level features without increasing the computational burden, we directly apply a convolution block to the larger-size input $I_{h}$ to obtain the feature $HR0$, without passing it through the backbone. 
The lowest-level feature $HR0$ with a size of 1024 is crucial for the enhanced structure perception in~\ourmodel\ (discussed in \secref{ablation}).

\subsection{Trunk and Structure Decoder}

Following the unite-divide-unite theme, after passing the union encoder, the two sets of recombined features are processed by the structure and trunk decoder respectively to fully exploit holistic trunk information and detailed structure information.
Inspired by~\cite{wei2020label}, we decouple the ground-truth label into the trunk and structure labels for bilateral supervision.

\subsubsection{Trunk Decoder}
As shown in~\figref{fig:model}, the features from deeper layers with 64 channels  $\{Ti|i=1, 2, 3, 4, 5\}$ are taken as input for the trunk decoder. 
Here, a dense cascade fusion strategy is applied to spotlight the trunk location and minimize the loss of spine features. 
Specifically, we perform progressive fusion at each upsampling stage.
For instance, the feature $T1$ is up-sampled by a factor of 2 and passed through a 1$\times$1 convolution block. Meanwhile, $T2$ is also processed through a 1$\times$1 convolution block and then summed with $T1$ to generate $T21$. 
We repeat this process in parallel operations to obtain $T31$, $T41$, and $T51$. 
After four upsampling dense fusions, we reach the unified trunk feature $T54$. 
Finally, the predicted trunk map is obtained by a 3$\times$3 convolution and a sigmoid function, with supervision using the trunk label. In addition, $T32$, $T43$, and $T54$ will be fed into the union decoder for incorporation with multiscale structure features.

\subsubsection{Structure Decoder}
Different from trunk feature fusion, mining structure information requires large-scale holistic features and specific low-level texture cues. 
To this end, the structure decoder receives additional large-scale coarse feature $HR0$ processed directly by a set of 1$\times$1 and 3$\times$3 convolutional blocks without passing through the backbone. 
To reduce the influence of smaller-scale input features $(LR1, LR2, LR3)$ on the identification of exterior details and hollow regions, we adopt a feature filtering operation to concentrate on interior and exterior edges and suppress undesired trunk noise. 
Given the smaller-size feature $LR3$ and the intermediate feature $T32$ from the trunk decoder with the same scale, we first reduce the feature $T32$ from 64 to 32 channels by 1$\times$1 convolution and BN, which is followed by a subtraction operation to obtain the filtered feature $S1$. This operation can be formulated as
\begin{equation}\label{equ:filter}
S1=LR3-\mathcal{C}_{1}(T32),
\end{equation}
where $\mathcal{C}_{1}$ is 1$\times$1 convolution with BN. Similarly, we can obtain the filtered features $S2$ and $S3$. As described at the top of~\figref{fig:model}, we employ a simple upsampling fusion to generate the integrated structure feature $S65$. Each step of the fusion is the same as the operation in the trunk decoder. Correspondingly, the structure label is used to supervise the structure map.

\subsection{Union Decoder}

\begin{figure}[t!]
\centering
\includegraphics[width=\linewidth]{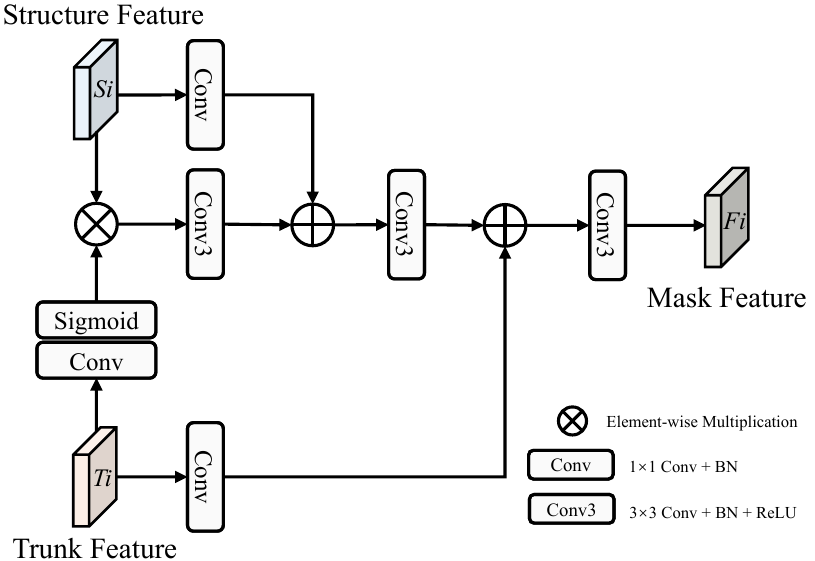}
\caption{Diagram of the trunk-structure aggregation module (TSA). The mask-structure aggregation module (MSA) shares the same architecture but replaces the trunk feature with the mask feature derived from the previous step in union decoder.}\label{fig:TSA}
\end{figure}

The union decoder aims to integrate multiscale cross-structural feature maps from both trunk and structure decoders. 
In this regard, we propose the trunk-structure aggregation module (TSA) and the mask-structure aggregation module (MSA) to combine trunk features and structure features of the same scale respectively, followed by layer-by-layer residual fusion to generate the unified feature.

The proposed TSA is developed to supplement detailed structural features while preserving sufficient trunk information. 
The workflow of TSA is illustrated in~\figref{fig:TSA}. The trunk feature $Ti$ is processed by 1$\times$1 convolution and a sigmoid function to produce the attention map. 
This trunk attention map is element-wise multiplied with the structural feature map to highlight border features and suppress background noise.
Then, we apply the residual operation separately to embed $Ti$ and $Si$ to enhance the coherence of the filtered features. 
After passing a 3$\times$3 convolution block, we attain the integrated mask feature $Fi$. 
The entire process of TSA can be formulated as follows:
\begin{equation}
Fi=\mathcal{C}_{3}(\mathcal{C}_{3}(\mathcal{C}_{3}(Si\otimes{Sig(\mathcal{C}_{1}(Ti))})+\mathcal{C}_{1}(Si))+\mathcal{C}_{1}(Ti)),
\end{equation}
where $\mathcal{C}_{3}$ is 3$\times$3 convolution followed by BN and ReLU. $Sig(\cdot)$ denotes the sigmoid function. 
Next, we upsample $Fi$ and add it to the mask feature $F(i+1)$ of the subsequent TSA output. 
When receiving $F3$, the unified feature already contains abundant trunk information. 

To further improve the accuracy of the target structure, we merge the mask feature directly with the structure feature using the MSA module at larger scales. 
Specifically, we replace the trunk feature input with the mask feature while keeping all other operations the same. 
Thanks to MSA, the noise from larger-scale structure features is further suppressed by the guidance of the mask feature. 
In parallel, the mask feature can be enriched with additional detailed information. 
Passing through three MSAs with residual operations, the union decoder yields the integrated mask feature $F$ and generates the final high-accuracy map by two 3$\times$3 convolution blocks.

\subsection{Loss Function}

According to the trunk label supervision, structure label supervision, and mask label supervision in our three decoders, the total loss function of~\ourmodel~can be defined as:
\begin{equation}
\mathcal{L}_{total}=\mathcal{L}_{trunk}+\mathcal{L}_{struc}+\mathcal{L}_{mask},
\end{equation}
where $L_{trunk}$ represents the loss of the trunk map in the trunk decoder and $L_{struc}$ indicates the loss of the structure map in the structure decoder.
Both $L_{trunk}$ and $L_{struc}$ are calculated using the Binary Cross Entropy (BCE) loss $L_{bce}$. 
Additionally, $L_{mask}$ is the loss of supervision on the final prediction map, which consists of the BCE loss and the Intersection over Union (IoU) loss~\cite{qin2019basnet}:
\begin{equation}
\mathcal{L}_{mask}=\mathcal{L}_{bce}+\mathcal{L}_{iou}.
\end{equation}
$L_{iou}$ is employed to enhance the supervision of the overall similarity of the prediction map, which is formulated as:
\begin{equation}
\mathcal{L}_{iou}=1-\frac{\sum_{(x,y)}f(x,y)g(x,y)}{\sum_{(x,y)}[f(x,y)+g(x,y)-f(x,y)g(x,y)]},
\end{equation}
where $f(x,y)$ denotes as the foreground probability of the predicted object and $g(x,y)$ is the ground-truth label.

\begin{table*}[t!]
\centering
	\caption{Quantitative comparison with 16 representative methods on the DIS5K dataset. $\uparrow$ / $\downarrow$ represents the higher/lower the score, the better. The best three scores are highlighted in \tr{red}, \tg{green}, and \tb{blue}, respectively. R is ResNet~\cite{he2016deep} and R2 is Res2Net~\cite{gao2019res2net}.
	}\label{tab:sota}
    \resizebox{\textwidth}{!}{
	\begin{tabular}{c|r|ccccccccccccccc|cc}
		\hline
        Dataset	&	Metric	&	        	
        \tabincell{c}{BASNet\\\cite{qin2019basnet}}	&	
         \tabincell{c}{GateNet\\\cite{zhao2020suppress}}	&	
        \tabincell{c}{U$^2$Net\\\cite{qin2020u2}}	&
        \tabincell{c}{HRNet\\\cite{wang2020deep}} &
         \tabincell{c}{LDF\\\cite{wei2020label}}	&	
        \tabincell{c}{F$^3$Net\\\cite{wei2020f3net}}	&	
        \tabincell{c}{GCPANet\\\cite{chen2020global}}	&
        \tabincell{c}{SINetV2\\\cite{fan2021concealed}}	&	
        \tabincell{c}{PFNet\\\cite{mei2021camouflaged}}	&	

        \tabincell{c}{MSFNet\\\cite{zhang2021auto}}	&	
        \tabincell{c}{CTDNet\\\cite{zhao2021complementary}}	&
        \tabincell{c}{BSANet\\\cite{zhu2022can}}	&

        \tabincell{c}{ISDNet\\\cite{guo2022isdnet}}	&	
        \tabincell{c}{IFA\\\cite{hu2022learning}}	&	
        \tabincell{c}{PGNet\\\cite{xie2022pyramid}}	&

       \tabincell{c}{IS-Net\\\cite{qin2022highly}}	&	
        \textbf{\ourmodel}	\\

% 注意zoomnet输入512  上采样1024  下采样256
        
        \hline 
        \multirow{4}{*}{\begin{sideways}\tabincell{c}{\textbf{Attr.}}\end{sideways}}

         &	Volume	&	CVPR19  & ECCV20 & PR20  & TPAMI20 & CVPR20 & AAAI20 & AAAI20& TPAMI21 & CVPR21 & MM21  & MM21 & AAAI22 & CVPR22 & ECCV22 & CVPR22 & ECCV22 & - \\

        &	Backbone &	R-50                        & R-50   & -     & -       & R-50   & R-50   & R-50                         & R2-50   & R-50   & R-50  & R-50                         & R2-50  & R-50   & R-50   & R-50                         & -                           & R-50                         \\

        &	Input Size	&	$1024^2$	&	$1024^2$	&	$1024^2$	&	$1024^2$	&	$1024^2$	&	$1024^2$ 	&	$1024^2$ &	$1024^2$		  &  $1024^2$    &$1024^2$    &$1024^2$&	$1024^2$   & $1024^2$	     &	$1024^2$&$1024^2$&	$1024^2$&$1024^2$	\\

        &	Size (MB)	&359.0                       & 515.0  & 176.3 & 264.4   & 101.0  & 102.6  & 268.7                        & 108.5   & 186.6  & 113.9 & 98.9                         & 131.1  & 111.5  & 111.4  & 150.1                        & 176.6                       & 100.2                        \\

        &	FPS    		&79.0                        & 76.7   & 50.0  & 24.7    & 72.6   & 61.0   & 59.0                         & 50.1    & 43.9   & 64.0  & 78.0                         & 35.0   & 78.6   & 33.7   & 64.3                         & 51.3                        & 45.5                         \\

\hline 
        
\multirow{6}{*}{\begin{sideways}\textbf{DIS-VD}\end{sideways}}

&	$maxF_\beta\uparrow$ &  0.737                       & 0.790  & 0.753 & 0.726   & 0.780  & 0.783  & {\cellcolor{myGreen!30} 0.798} & 0.748   & 0.793  & 0.714 & {\cellcolor{myBlue!30} 0.795} & 0.738  & 0.763  & 0.749  & {\cellcolor{myGreen!30} 0.798} & 0.791                       & {\cellcolor{myRed!30} 0.823} \\
&	$F^w_\beta\uparrow$ &0.656                       & 0.716  & 0.656 & 0.641   & 0.715  & 0.714  & {\cellcolor{myGreen!30} 0.736} & 0.632   & 0.724  & 0.620 & 0.729                        & 0.615  & 0.691  & 0.653  & {\cellcolor{myBlue!30} 0.733} & 0.717                       & {\cellcolor{myRed!30} 0.763} \\
&	$~M~\downarrow$ &0.094                       & 0.072  & 0.089 & 0.095   & 0.071  & 0.072  & {\cellcolor{myGreen!30} 0.066} & 0.093   & 0.075  & 0.105 & {\cellcolor{myBlue!30} 0.067} & 0.100  & 0.080  & 0.088  & {\cellcolor{myBlue!30} 0.067} & 0.074                       & {\cellcolor{myRed!30} 0.059} \\
&	$S_{\alpha}\uparrow$&0.781                       & 0.816  & 0.785 & 0.767   & 0.813  & 0.813  & {\cellcolor{myGreen!30} 0.826} & 0.796   & 0.818  & 0.759 & 0.823                        & 0.786  & 0.803  & 0.785  & {\cellcolor{myBlue!30} 0.824} & 0.813                       & {\cellcolor{myRed!30} 0.838} \\
&	$E_{\phi}^{m}\uparrow$&0.809                       & 0.868  & 0.809 & 0.824   & 0.861  & 0.860  & {\cellcolor{myBlue!30} 0.878} & 0.814   & 0.868  & 0.800 & 0.873                        & 0.807  & 0.852  & 0.829  & {\cellcolor{myGreen!30} 0.879} & 0.856                       & {\cellcolor{myRed!30} 0.892} \\
&	$HCE_\gamma\downarrow$&{\cellcolor{myBlue!30} 1132 }                      & 1189   & 1139  & 1 560   & 1314   & 1291   & 1264                         & 1756    & 1350   & 1376  & 1354                         & 1731   & 1549   & 1437   & 1326                         & {\cellcolor{myGreen!30} 1116} & {\cellcolor{myRed!30} 1097}  \\

	\hline 
\multirow{6}{*}{\begin{sideways}\textbf{DIS-TE1}\end{sideways}}

&	$maxF_\beta\uparrow$ &  0.663                       & 0.737  & 0.701 & 0.668   & 0.727  & 0.726  & {\cellcolor{myBlue!30} 0.741} & 0.695   & 0.740  & 0.658 & 0.738                        & 0.683  & 0.717  & 0.673  & {\cellcolor{myGreen!30} 0.754} & 0.740                       & {\cellcolor{myRed!30} 0.784} \\
&	$F^w_\beta\uparrow$ &0.577                       & 0.654  & 0.601 & 0.579   & 0.655  & 0.655  & {\cellcolor{myBlue!30} 0.676} & 0.568   & 0.665  & 0.559 & 0.668                        & 0.545  & 0.643  & 0.573  & {\cellcolor{myGreen!30} 0.680} & 0.662                       & {\cellcolor{myRed!30} 0.720} \\
&	$~M~\downarrow$ &0.105                       & 0.076  & 0.085 & 0.088   & 0.074  & 0.076  & {\cellcolor{myBlue!30} 0.070} & 0.088   & 0.075  & 0.101 & 0.072                        & 0.098  & 0.077  & 0.088  & {\cellcolor{myGreen!30}0.067} & 0.074                       & {\cellcolor{myRed!30} 0.059} \\
&	$S_{\alpha}\uparrow$&0.741                       & 0.786  & 0.762 & 0.742   & 0.783  & 0.783  & {\cellcolor{myBlue!30} 0.797} & 0.767   & 0.791  & 0.734 & 0.792                        & 0.754  & 0.782  & 0.746  & {\cellcolor{myGreen!30} 0.800} & 0.787                       & {\cellcolor{myRed!30} 0.817} \\
&	$E_{\phi}^{m}\uparrow$&0.756                       & 0.826  & 0.783 & 0.797   & 0.822  & 0.820  & 0.834                        & 0.785   & 0.830  & 0.771 & {\cellcolor{myBlue!30} 0.837} & 0.773  & 0.824  & 0.785  & {\cellcolor{myGreen!30} 0.848} & 0.820                       & {\cellcolor{myRed!30} 0.864} \\
&	$HCE_\gamma\downarrow$&155                         & 164    & 165   & 262     & 155    & 150    & {\cellcolor{myGreen!30} 145}   & 320     & 164    & 203   & 162                          & 314    & 214    & 229    & 162                          & {\cellcolor{myBlue!30} 149}  & {\cellcolor{myRed!30} 140}   \\

	\hline 
\multirow{6}{*}{\begin{sideways}\textbf{DIS-TE2}\end{sideways}}	

&	$maxF_\beta\uparrow$ &  0.738                       & 0.795  & 0.768 & 0.747   & 0.784  & 0.789  & {\cellcolor{myBlue!30} 0.799} & 0.761   & 0.796  & 0.736 & {\cellcolor{myBlue!30} 0.799} & 0.752  & 0.783  & 0.758  & {\cellcolor{myGreen!30} 0.807} & 0.799                       & {\cellcolor{myRed!30} 0.829} \\
&	$F^w_\beta\uparrow$ &0.653                       & 0.721  & 0.676 & 0.664   & 0.719  & 0.719  & {\cellcolor{myBlue!30} 0.741} & 0.646   & 0.729  & 0.642 & 0.731                        & 0.628  & 0.714  & 0.666  & {\cellcolor{myGreen!30} 0.743} & 0.728                       & {\cellcolor{myRed!30} 0.768} \\
&	$~M~\downarrow$ &0.096                       & 0.074  & 0.083 & 0.087   & 0.073  & 0.075  & {\cellcolor{myBlue!30} 0.068} & 0.090   & 0.073  & 0.096 & 0.070                        & 0.098  & 0.072  & 0.085  & {\cellcolor{myGreen!30} 0.065} & 0.070                       & {\cellcolor{myRed!30} 0.058} \\
&	$S_{\alpha}\uparrow$&0.781                       & 0.818  & 0.798 & 0.784   & 0.813  & 0.814  & {\cellcolor{myBlue!30} 0.830} & 0.805   & 0.821  & 0.780 & 0.823                        & 0.794  & 0.817  & 0.793  & {\cellcolor{myGreen!30} 0.833} & 0.823                       & {\cellcolor{myRed!30} 0.843} \\
&	$E_{\phi}^{m}\uparrow$&0.808                       & 0.864  & 0.825 & 0.840   & 0.862  & 0.860  & {\cellcolor{myBlue!30} 0.874} & 0.823   & 0.866  & 0.816 & 0.872                        & 0.815  & 0.865  & 0.835  & {\cellcolor{myGreen!30} 0.880} & 0.858                       & {\cellcolor{myRed!30} 0.886} \\
&	$HCE_\gamma\downarrow$&{\cellcolor{myBlue!30} 341}  & 368    & 367   & 555     & 370    & 358    & 345                          & 672     & 389    & 456   & 382                          & 660    & 494    & 479    & 375                          & {\cellcolor{myGreen!30} 340}  & {\cellcolor{myRed!30} 325}   \\

    \hline 
\multirow{6}{*}{\begin{sideways}\textbf{DIS-TE3}\end{sideways}}	

&	$maxF_\beta\uparrow$ &  0.790                       & 0.835  & 0.813 & 0.784   & 0.828  & 0.824  & {\cellcolor{myGreen!30} 0.844} & 0.791   & 0.835  & 0.763 & 0.838                        & 0.783  & 0.817  & 0.797  & {\cellcolor{myBlue!30} 0.843} & 0.830                       & {\cellcolor{myRed!30} 0.865} \\
&	$F^w_\beta\uparrow$ &0.714                       & 0.769  & 0.721 & 0.700   & 0.770  & 0.762  & {\cellcolor{myGreen!30} 0.789} & 0.676   & 0.771  & 0.674 & 0.778                        & 0.660  & 0.747  & 0.705  & {\cellcolor{myBlue!30} 0.785} & 0.758                       & {\cellcolor{myRed!30} 0.809} \\
&	$~M~\downarrow$ &0.080                       & 0.062  & 0.073 & 0.080   & 0.061  & 0.063  & {\cellcolor{myGreen!30} 0.056} & 0.084   & 0.062  & 0.089 & {\cellcolor{myBlue!30} 0.059} & 0.090  & 0.065  & 0.077  & {\cellcolor{myGreen!30} 0.056} & 0.064                       & {\cellcolor{myRed!30} 0.050} \\
&	$S_{\alpha}\uparrow$&0.816                       & 0.847  & 0.823 & 0.805   & 0.844  & 0.841  & {\cellcolor{myGreen!30} 0.855} & 0.823   & 0.847  & 0.793 & {\cellcolor{myBlue!30} 0.850} & 0.814  & 0.834  & 0.815  & 0.844                        & 0.836                       & {\cellcolor{myRed!30} 0.865} \\
&	$E_{\phi}^{m}\uparrow$&0.848                       & 0.901  & 0.856 & 0.869   & 0.896  & 0.892  & {\cellcolor{myBlue!30} 0.909} & 0.845   & 0.901  & 0.845 & 0.903                        & 0.840  & 0.893  & 0.861  & {\cellcolor{myGreen!30}0.911} & 0.883                       & {\cellcolor{myRed!30} 0.917} \\
&	$HCE_\gamma\downarrow$&{\cellcolor{myGreen!30} 681}  & 737    & 738   & 1049    & 782    & 767    & 742                          & 1219    & 816    & 901   & 807                          & 1204   & 994    & 937    & 797                          & {\cellcolor{myBlue!30} 687}  & {\cellcolor{myRed!30} 658}   \\

	\hline 
 
\multirow{6}{*}{\begin{sideways}\textbf{DIS-TE4}\end{sideways}}	

&	$maxF_\beta\uparrow$ &  0.785                       & 0.826  & 0.800 & 0.772   & 0.818  & 0.815  & {\cellcolor{myGreen!30} 0.831} & 0.763   & 0.816  & 0.743 & 0.826                        & 0.757  & 0.794  & 0.790  & {\cellcolor{myGreen!30} 0.831} &   {\cellcolor{myBlue!30} 0.827}                     & {\cellcolor{myRed!30} 0.846} \\
&	$F^w_\beta\uparrow$ &0.713                       & 0.766  & 0.707 & 0.687   & 0.762  & 0.753  & {\cellcolor{myGreen!30} 0.776} & 0.649   & 0.755  & 0.660 & 0.766                        & 0.640  & 0.725  & 0.700  & {\cellcolor{myBlue!30} 0.774} & 0.753                       & {\cellcolor{myRed!30} 0.792} \\
&	$~M~\downarrow$ &0.087                       & 0.067  & 0.085 & 0.092   & 0.067  & 0.070  & {\cellcolor{myGreen!30} 0.064} & 0.101   & 0.072  & 0.102 & 0.066                        & 0.107  & 0.079  & 0.085  & {\cellcolor{myBlue!30} 0.065} & 0.072                       & {\cellcolor{myRed!30} 0.059} \\
&	$S_{\alpha}\uparrow$&0.806                       & 0.839  & 0.814 & 0.792   & 0.832  & 0.826  & {\cellcolor{myGreen!30} 0.841} & 0.799   & 0.831  & 0.775 & {\cellcolor{myBlue!30} 0.840} & 0.794  & 0.815  & 0.811  & {\cellcolor{myGreen!30} 0.841} & 0.830                       & {\cellcolor{myRed!30} 0.849} \\
&	$E_{\phi}^{m}\uparrow$&0.844                       & 0.895  & 0.837 & 0.854   & 0.888  & 0.883  &{\cellcolor{myBlue!30}  0.898}                        & 0.816   & 0.885  & 0.825 & 0.895                        & 0.815  & 0.873  & 0.847  & {\cellcolor{myGreen!30} 0.899} & 0.870                       & {\cellcolor{myRed!30} 0.901} \\
&	$HCE_\gamma\downarrow$&{\cellcolor{myGreen!30} 2852} & 2965   & 2898  & 3864    & 3364   & 3291   & 3229                         & 4050    & 3391   & 3425  & 3447                         & 4014   & 3760   & 3554   & 3361                         & {\cellcolor{myBlue!30} 2888} & {\cellcolor{myRed!30} 2785}  \\

        \hline 
	\hline
\multirow{6}{*}{\begin{sideways}\tabincell{c}{\textbf{Overall}\\\textbf{DIS-TE (1-4)}}\end{sideways}}

&	$maxF_\beta\uparrow$ &  0.744                       & 0.798  & 0.771 & 0.743   & 0.789  & 0.789  & {\cellcolor{myBlue!30} 0.804} & 0.753   & 0.797  & 0.725 & 0.800                        & 0.744  & 0.778  & 0.755  & {\cellcolor{myGreen!30} 0.809} & 0.799                       & {\cellcolor{myRed!30} 0.831} \\
&	$F^w_\beta\uparrow$ &0.664                       & 0.728  & 0.676 & 0.658   & 0.727  & 0.722  & {\cellcolor{myGreen!30} 0.746} & 0.635   & 0.730  & 0.634 & {\cellcolor{myBlue!30} 0.736} & 0.618  & 0.707  & 0.661  & {\cellcolor{myGreen!30} 0.746} & 0.726                       & {\cellcolor{myRed!30} 0.772} \\
&	$~M~\downarrow$ &0.092                       & 0.070  & 0.082 & 0.087   & 0.069  & 0.071  & {\cellcolor{myBlue!30} 0.065} & 0.091   & 0.071  & 0.097 & 0.067                        & 0.098  & 0.073  & 0.084  & {\cellcolor{myGreen!30} 0.063} & 0.070                       & {\cellcolor{myRed!30} 0.057} \\
&	$S_{\alpha}\uparrow$&0.786                       & 0.823  & 0.799 & 0.781   & 0.818  & 0.816  & {\cellcolor{myGreen!30} 0.831} & 0.799   & 0.823  & 0.771 & 0.826                        & 0.789  & 0.812  & 0.791  & {\cellcolor{myBlue!30} 0.830} & 0.819                       & {\cellcolor{myRed!30} 0.844} \\
&	$E_{\phi}^{m}\uparrow$&0.814                       & 0.872  & 0.825 & 0.840   & 0.867  & 0.864  & {\cellcolor{myBlue!30} 0.879} & 0.817   & 0.871  & 0.814 & 0.877                        & 0.811  & 0.864  & 0.832  & {\cellcolor{myGreen!30}0.885} & 0.858                       & {\cellcolor{myRed!30} 0.892} \\
&	$HCE_\gamma\downarrow$&{\cellcolor{myGreen!30} 1007} & 1059   & 1042  & 1432    & 1167   & 1141   & 1115                         & 1565    & 1190   & 1246  & 1200                         & 1548   & 1365   & 1299   & 1173                         & {\cellcolor{myBlue!30}1016} & {\cellcolor{myRed!30} 977}  \\

        \hline
        
	\end{tabular}
	}
\end{table*}

% &\cellcolor{myBlue!30} 1009      &\cellcolor{myGreen!30} 999   &\cellcolor{myRed!30}977  \\

\section{Experiments}

\subsection{Datasets and Evaluation Metrics}

\subsubsection{Datasets}  
%DIS5K Datasets.   DIS5K有5470张图像，包好三个子集：dis tr（3000）、DIS-VD（470）和DIS-TE（2000），分别用于训练、验证和测试。根据数据集的对象形状和结构复杂性，2000张DIS-TE图像被进一步分为四个子集（DIS-TE1∼DIS-TE4），形状复杂性（结构复杂度和边界复杂性的乘法）递增，每个子集中有500张图像，代表四个测试难度级别。
%The DIS5K dataset consists of 5470 images, which are divided into three subsets: DIS-TR (3000 images) for training, DIS-VD (470 images) for validation, and DIS-TE (2000 images) for testing. Based on the shape and structural complexity of the objects in the dataset, the 2000 images in DIS-TE are further divided into four subsets (DIS-TE1 to DIS-TE4), with increasing shape complexity (multiplication of structural complexity and boundary complexity). Each subset contains 500 images, representing four levels of testing difficulty.
We conduct experiments on DIS5K~\cite{qin2022highly} with a total of 5,470 images, which are divided into three subsets: DIS-TR (3,000 images for training), DIS-VD (470 images for validation), and DIS-TE (2,000 images for testing). 
Depending on the shape and structural complexity of the objects, DIS-TE are further divided into four subsets (DIS-TE1 to DIS-TE4). 
Each subset contains 500 images, representing four levels of testing difficulty.
 
 %为了从不同角度评估高精度二义分割的性能，我们采用五种广泛使用的指标以及HCE来评估所有比较方法： 包括最大↑)[2]，测量（Fβ mx加权↑）[2]，测量（Fβ↑）[64]，平均绝对误差（M↓）[73]，结构测量（Sα↑）[22]，意味着增强对齐测量（Eφ↑）[23,25]和人工矫正量（HCEγ↓），作为前面5个指标的补充，HCE近似于纠正错误预测所需的人工努力，以满足现实应用中的特定精度要求，它对目标的细节和整体性非常敏感，其中γ代表误差容忍度（表示被忽略的小故障区域的大小），我们设置γ = 5和[xx]保持一致 

\subsubsection{Evaluation Metrics}  
% In order to evaluate the performance of high- accuracy Dichotomous segmentation from different perspectives, we adopt five commonly used metrics as well as HCE (Human Correction Efforts) to evaluate all compared methods: including maximal F-measure ($F_\beta^\text{max}\uparrow$) [2], weighted F-measure ($F_\beta^\omega \uparrow$) [64], mean absolute error $(M\downarrow$) [73], structural measure ($S_\alpha \uparrow$ ) [22], mean enhanced alignment measure ($E_{\phi}^\text{m} \uparrow$) [23, 25] and human correction efforts ($HCE_{\gamma} \downarrow$). As a complement to the previous five metrics, HCE approximates the human effort required to correct prediction errors in order to meet specific accuracy requirements in real-world applications. It is highly sensitive to the details and integrity of the object. Here, $\gamma$ represents the error tolerance (indicating the size of small error regions to be ignored), and we set $\gamma$ = 0.5 to be consistent with [xx].
To comprehensively assess the performance of high-accuracy DIS, we employ a total of six evaluation metrics to evaluate compared models in terms of both foreground region accuracy and structural granularity.
To assess the accuracy of foreground areas, we adopt five metrics that are widely used in category-agnostic segmentation tasks~\cite{wang2021salient, fan2021concealed}. These metrics include maximal F-measure ($F_\beta^\text{max}\uparrow$)~\cite{achanta2009frequency}, weighted F-measure ($F_\beta^\omega\uparrow$)~\cite{margolin2014evaluate}, Mean Absolute Error (MAE, $M\downarrow$)~\cite{perazzi2012saliency}, Structural measure (S-measure, $S_\alpha\uparrow$)~\cite{fan2017structure}, and mean Enhanced alignment measure (E-measure, $E_{\phi}^\text{m}\uparrow$)~\cite{fan2018enhanced}. 
For further evaluating
 the quality of structural details and integrity, 
following~\cite{qin2022highly}, we also utilize Human Correction Efforts (HCE$_{\gamma}\downarrow$) to specifically focus on the details and integrity of the object. 
Here, $\gamma$ represents the error tolerance, which is set to 5 to maintain consistent with~\cite{qin2022highly}.

\subsection{Implementation Details} 

Experiments are implemented in PyTorch on a single RTX 3090 GPU. 
During the training phase, the images are resized to both 1024$\times$1024 and 256$\times$256 to create the dual-size input. 
Horizontal flipping and random cropping are applied for data augmentation. 
The ResNet~\cite{he2016deep} is used as the backbone with the pre-trained weights on ImageNet~\cite{deng2009imagenet}, while other parameters of~\ourmodel~are initialized randomly. 
Notably, the low-level feature at a scale of 1024 is processed directly by a set of convolutional operations without any pre-trained weights. 
The maximum learning rate for the backbone is set to 0.005, while other parts are set to 0.05. 
To optimize the training process, we utilize the Stochastic gradient descent (SGD) optimizer with a warm-up strategy and linear decay strategy. 
The batch size is set to 8, and the maximum number of epochs is set to 48. 
During inference, the input image with a scale of 1024$\times$1024 is automatically resized to 256$\times$256 to form a double-size input, which is then fed into the network for prediction without any post-processing.

\subsection{Comparison with State-of-the-arts}
%这里说后续加的几个sota 能处理高分辨率的都处理成1024 r50?

% 我们和18个不同....

% 值得注意的是，我们在~\cite{qin2022highly}establish DIS benchmark 基础上，新增加了最新的几个SOTA，比如LDF~\cite{wei2020label}(CVPR2020),PGNet~\cite{xie2022pyramid}(CVPR2022)
% Zoom-Net~\cite{pang2022zoom}(CVPR2022)
% IFA~\cite{hu2022learning}(ECCV2022) and ISDNet~\cite{guo2022isdnet}(CVPR2022),为了公平比较，对于如果可以直接处理高分辨率输入的模型，我们将模型最大输入尺度统一为1024，对于多encoder 网络最重的backbone使用resnet50 。

% Tab 1 shows the quantitative results.AS we can seen,
% 我们的方法在六个评价指标上产生最佳的性能，特别的，对比专门为DIS任务设计的ISnet 我们的三版本(with不同backbone)的模型在overall DIS-TE(1-4) 都取得了更好的性能，其中使用Resnet18的那个ones,模型大小只有49MB，而ISNet达到了176.6MB.
% 关键的是，从表格中可以发现以往的模型在前5个指标和HCE指标的性能冲突，往往不能同时提升，而我们的模型Joint Boosting Structure and Trunk精度，使得所有指标都有一个显著提升。
% 从表中还可以看出，ZOOMNET和PGNet在S-measure 也上展示出比较优异的精度，这验证了多尺度输入对结构完整度的必要性。对比他们，我们采用union encoder的方式可以更全面提升所有指标精度，以更少的参数。
\subsubsection{Quantitative Comparisons}
%Notably, based on the DIS benchmark established in~\cite{qin2022highly}, we have added several latest SOTA models, such as LDF~\cite{wei2020label} (CVPR2020), PGNet~\cite{xie2022pyramid} (CVPR2022), ZoomNet~\cite{pang2022zoom} (CVPR2022), IFA~\cite{hu2022learning} (ECCV2022) and ISDNet~\cite{guo2022isdnet} (CVPR2022). 
%For fair comparison, if the models were capable of directly processing high-resolution inputs, we standardized the maximum input size of the models to 1024 $\times$1024. For multi-encoder networks, we used ResNet50 as the heaviest backbone.
We compare the proposed \ourmodel~with the DIS-only model IS-Net~\cite{qin2022highly} and other 15 well-known task-related models, including BASNet~\cite{qin2019basnet}, GateNet~\cite{zhao2020suppress}, $\rm U^\text{2}$Net~\cite{qin2020u2}, HRNet~\cite{wang2020deep}, LDF~\cite{wei2020label}, $\rm F^\text{3}Net$~\cite{wei2020f3net}, GCPANet~\cite{chen2020global}, SINet-V2~\cite{fan2021concealed}, PFNet~\cite{mei2021camouflaged}, MSFNet~\cite{zhang2021auto}, CTDNet~\cite{zhao2021complementary},   BSANet~\cite{zhu2022can}, ISDNet~\cite{guo2022isdnet}, IFA~\cite{hu2022learning}, and PGNet~\cite{xie2022pyramid}.
For a fair comparison, we standardize the input size of the comparison models to 1024$\times$1024. 
Besides, most models use the ResNet-50~\cite{he2016deep} as the default backbone with ImageNet~\cite{deng2009imagenet} pre-trained weights except for those using Res2Net-50~\cite{gao2019res2net} (\eg, SINetV2 and BSANet) and random initialization (\eg, IS-Net, $\rm U^\text{2}$-$Net$, and HRNet). 
% All models are trained on the DIS-TR set and evaluated on the sets of DIS-VD, DIS-TE1, DIS-TE2, DIS-TE3, DIS-TE4, and overall DIS-TE.

The comparison results are shown in the \tabref{tab:sota}. We can explicitly observe that our UDUN outperforms the compared models by a large margin across all test sets. In particular, we gain significant improvements in five general metrics compared to the second-place performance on the overall DIS-TE, \ie, 2.2\% on the $F_\beta^\text{max}$, 2.6\% on the $F_\beta^\omega$, 0.6\% on the $M$, 1.3\% on the $S_\alpha$, and 0.7\% on the $E_{\phi}^\text{m}$. This achievement is mainly attributed to the proposed unite-divide-unite strategy for precise identification of the integrity of targets. More importantly, for the HCE metric, focusing on a detailed structure assessment of predictions, our model also outperforms the current best performer, BASNet, by about 30 points. It demonstrates the ample extraction of structure information in our structure decoder and the effective integration in our union decoder.

\subsubsection{Qualitative Comparisons}
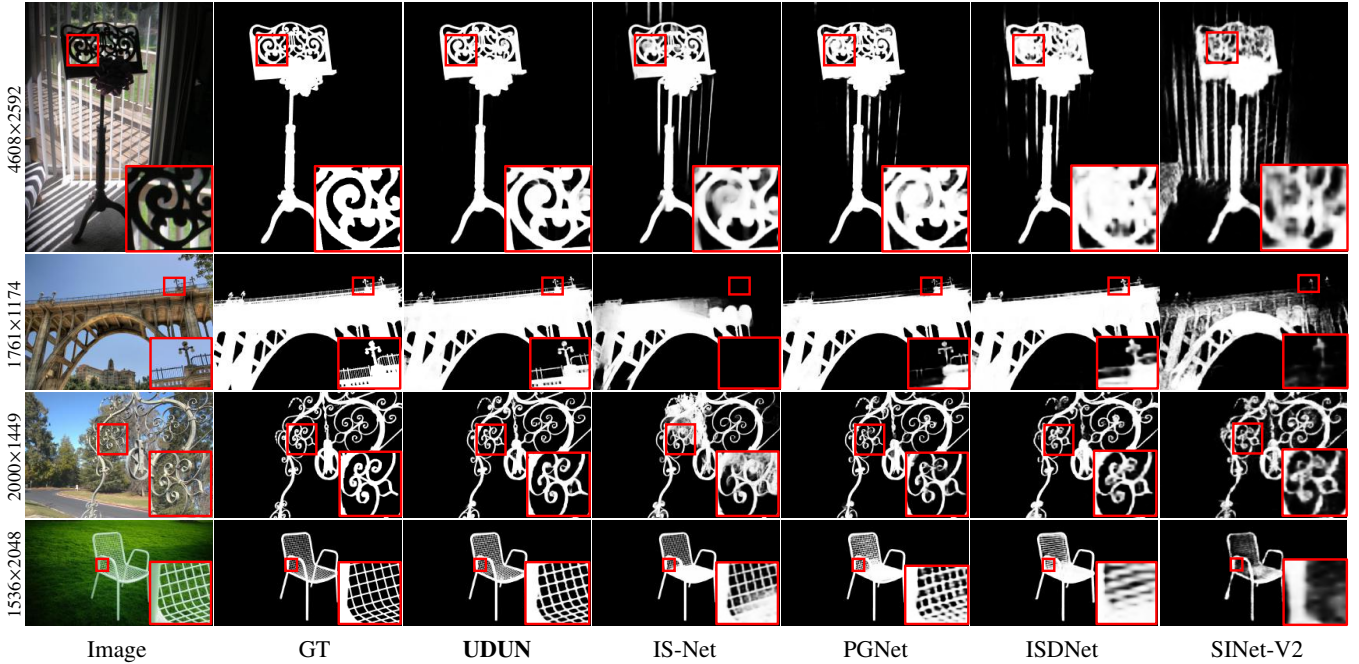
\begin{figure*}[t!]
\centering
\includegraphics[width=\textwidth]{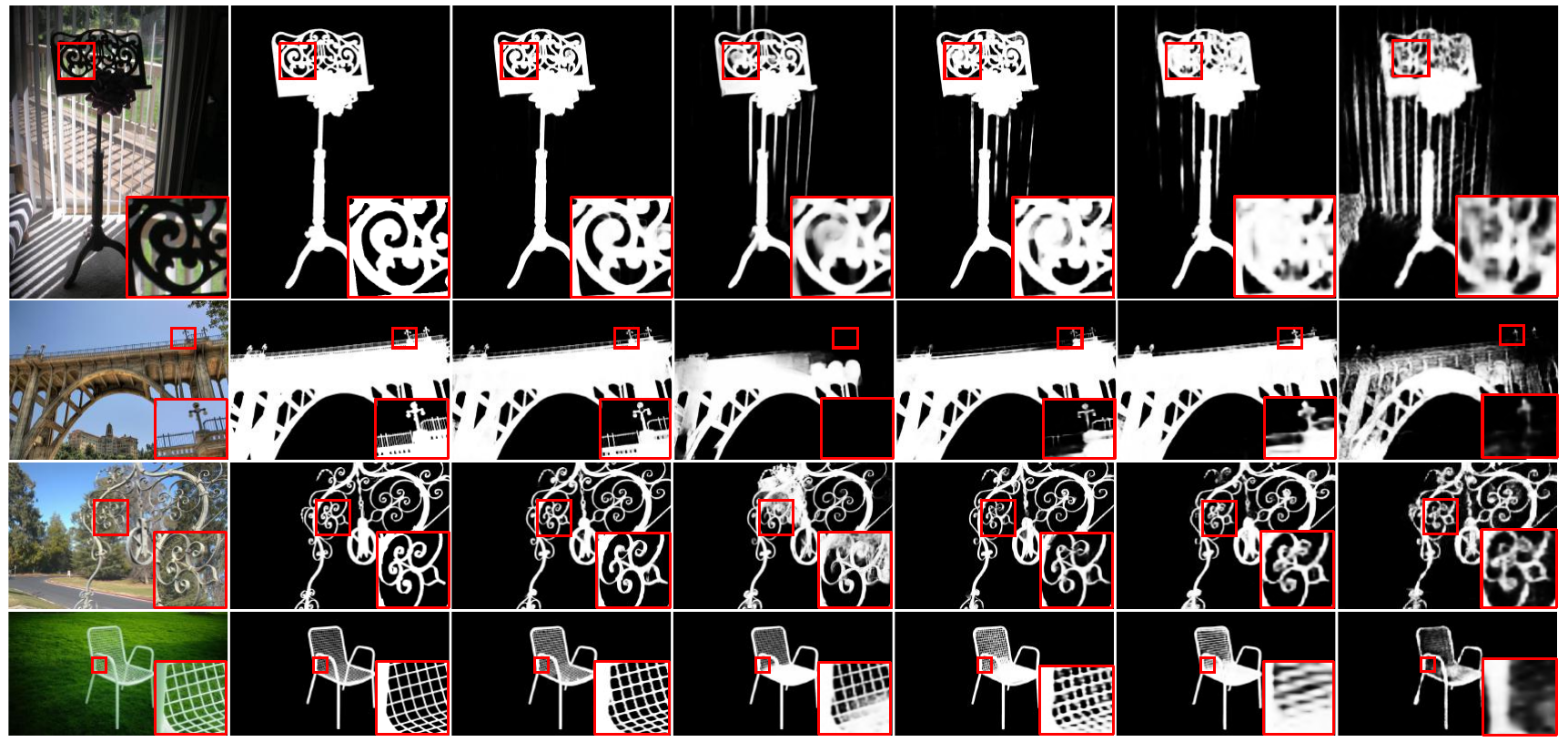}
    \put(-510,4.5){\rotatebox{90}{\footnotesize{1536$\times$2048}
}}
    \put(-510,49){\rotatebox{90}{\footnotesize{2000$\times$1449
}}}
    \put(-510,99){\rotatebox{90}{\footnotesize{1761$\times$1174
}}}
    \put(-510,173){\rotatebox{90}{\footnotesize{4608$\times$2592
}}}
    \put(-480,-10){{Image}}
    \put(-400,-10){{GT}}
    \put(-338,-10){\textbf{UDUN}}
    \put(-266,-10){{IS-Net}}
    \put(-194,-10){{PGNet}}
    \put(-125,-10){{ISDNet}}
    \put(-55,-10){SINet-V2}

\caption{Qualitative comparisons of our~\ourmodel~and other representative methods on overall DIS-TE.}\label{fig:visual_comp}
\end{figure*}

%得挑一下可视化结果，各个sota的  可以用python脚本GPT is ok     展示FPFN 区域  统计HCE（FP FN）
% image  GT our  IS-Net  PGnet   IFAnet  ISDnet
% 我们的模型不仅可以预测出trunk 区域  还可以保持结构的完整性，特别是interior structure，如图中的椅子....啥的
\figref{fig:visual_comp} exhibits visual comparisons of our method with four cutting-edge methods. 
On a macroscopic level, \ourmodel~provides a more complete segmentation of high-resolution targets than other competitors. 
On a microscopic level, our model is capable of handling complex structures and slender areas with higher accuracy. 
More visualization results can be found in the \sm{supplementary material}. 

% \begin{figure*}[t!]
% \centering
% \includegraphics[width=\textwidth]{Imgs/Qualitative_Cmp.pdf}

% \caption{Qualitative comparisons of our~\ourmodel~and other representative methods on overall DIS-TE.}\label{fig:visual_comp}
% \end{figure*}

% \begin{figure*}[t!]
% \centering
% \begin{overpic}[width=\textwidth]{}
%     \put(0.5,63){\rotatebox{90}{\small{COME-N}}}
%     \put(0.5,44){\rotatebox{90}{\small{COME-D}}}
%     \put(0.5,30){\rotatebox{90}{\small{SIP}}}
%     \put(0.5,10){\rotatebox{90}{\small{DSIS}}}
    
%     \put(16,0){\small{RGB}}
%     \put(45,0){\small{Object-GT}}
%     \put(74,0){\small{Instance-GT}}
% \end{overpic}
% \caption{Comparison of the consistency between salient object-level ground truth and instance-level ground truth in RGB-D saliency datasets.}\label{inconsistency_come15k}
% \end{figure*}

\subsection{Ablation Studies}\label{ablation}
%我们在DIS-VD数据集上评估评估每个提出的组件。采用Fβ max，MAE，HCE作为主要指标。结果分析如下。
To evaluate the effectiveness of each design of our~\ourmodel, we conduct a range of ablations on the DIS-VD validation set.

\subsubsection{Effectiveness of DCM}
The role of DCM is to divide and reassemble the two groups of multiscale cross-level features produced by the shared backbone. 
The decoupled higher-level features with lower sizes are suitable for our trunk decoder, whereas the lower-level features with higher sizes are appropriate for our structure decoder. 
% Referring to \tabref{tab:dcm}, regrouping features through DCM results in effective refinement and integration of decoupled features, thereby enhancing the accuracy of prediction maps.
Referring to \tabref{tab:dcm}, the regrouping of features by DCM effectively improves the capabilities of subsequent decoders.

\begin{table}
\footnotesize
  \renewcommand{\arraystretch}{1}
  \renewcommand{\tabcolsep}{2.8mm}
  \caption{Effectiveness of DCM in union encoder.}
  \label{tab:dcm}
  \begin{tabular}{c|cccccc}
    \toprule
    DCM & $F_\beta^\text{max} \uparrow$ &  $F_\beta^\omega \uparrow$ & $M \downarrow$ &
    $S_\alpha \uparrow$ & $E_{\phi}^\text{m} \uparrow$ & $HCE_{\gamma} \downarrow$ \\
    \midrule
       &0.815 & 0.753 & 0.064 & 0.833 & 0.886 & 1112\\
    \rowcolor[RGB]{235,235,235}
     \checkmark &\textbf{0.823} & \textbf{0.763} & \textbf{0.059} & \textbf{0.838} & \textbf{0.892} & \textbf{1097} \\
  \bottomrule
\end{tabular}
\end{table}

%使用HR作为structural  feature    LR作为Trunk feature  作为对比

\subsubsection{Trunk and Structure Decoders}
We conduct an ablation study to assess the effectiveness of the proposed trunk and structure decoders of our~\ourmodel~in \tabref{tab:decoder}. 
It should be noted that, in this experiment, we directly upsample and sum all multiscale features instead of the corresponding decoder that is removed. 
As demonstrated in \tabref{tab:decoder}, both the trunk and the structure decoders have a remarkable contribution to the segmentation results. When embedding two decoders simultaneously, our model achieves superior performance. 
To further intuitively illustrate the capability of trunk and structure decoders for refining and integrating features, we also provide the last layer of features from the corresponding decoders in \figref{fig:cam}. 
As displayed in \figref{fig:cam}(b) and \figref{fig:cam}(c), the trunk areas are identified by the trunk decoder, while the structure decoder concentrates on detecting detailed object structures.

\begin{table}
\footnotesize
  \renewcommand{\arraystretch}{1}
  \renewcommand{\tabcolsep}{1.9mm}
  \caption{Ablations for trunk and structure decoders.}
  \label{tab:decoder}
  \begin{tabular}{c|c|cccccc}
    \toprule
    Structure & Trunk & $F_\beta^\text{max} \uparrow$ &  $F_\beta^\omega \uparrow$ & $M \downarrow$ &
    $S_\alpha \uparrow$ & $E_{\phi}^\text{m} \uparrow$ & $HCE_{\gamma} \downarrow$ \\
    \midrule
     \checkmark  &  & 0.807 & 0.743 & 0.064 & 0.826 & 0.880  & 1110 \\
       & \checkmark & 0.812 & 0.752 & 0.062 & 0.833 & 0.886 & 1136 \\
    \rowcolor[RGB]{235,235,235}
    \checkmark  &  \checkmark & \textbf{0.823} & \textbf{0.763} & \textbf{0.059} & \textbf{0.838} & \textbf{0.892} & \textbf{1097} \\
  \bottomrule
\end{tabular}
\end{table}

\subsubsection{Effectiveness of TSA}
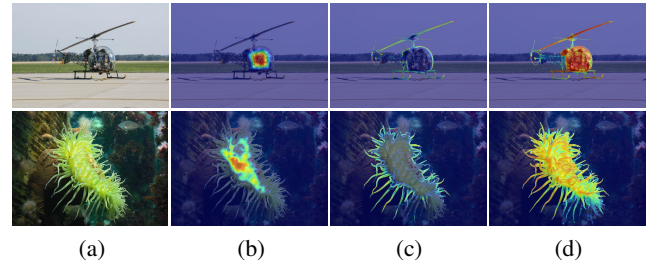
\begin{figure}[t!]
\centering
\includegraphics[width=\linewidth]{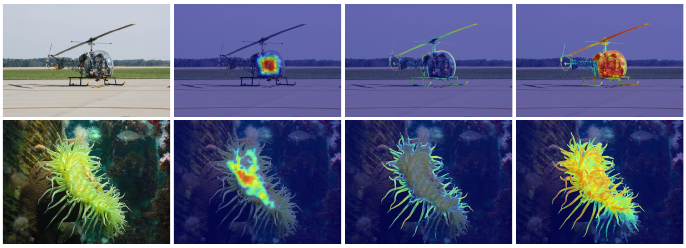}
  \put(-215,-10){{(a)}}
    \put(-155,-10){{(b)}}
    \put(-95,-10){{(c)}}
    \put(-35,-10){(d)}

\caption{Visualization of the output feature maps from three decoders. (a) Input image; (b) Trunk feature map $T54$; (c) Structure feature map $S65$; (d) Mask feature map $F$.}\label{fig:cam}
\end{figure}

The proposed TSA module is intended to combine multiscale cross-structure features from the trunk and structure decoders. 
In \tabref{tab:TSA}, we compare our TSA with the summation and concatenation operations.
The results show that it is not desirable to directly add two spatially complementary features, and adopting concatenation performs better than the summation operation.
Compared to summation and concatenation, TSA adequately integrates the trunk and structure information to generate a unified mask feature. 
This suggests that TSA suppresses the background noise of structure features via trunk features and enhances the the representation of body and detail information. 
\figref{fig:cam}(d) also visualizes the fused mask feature generated by  the union decoder.

\begin{table}
\footnotesize
  \renewcommand{\arraystretch}{1}
  \renewcommand{\tabcolsep}{2.6mm}
  \caption{Effectiveness of TSA in union decoder.}
  \label{tab:TSA}
  \begin{tabular}{c|cccccc}
    \toprule
    Modules & $F_\beta^\text{max} \uparrow$ &  $F_\beta^\omega \uparrow$ & $M \downarrow$ &
    $S_\alpha \uparrow$ & $E_{\phi}^\text{m} \uparrow$ & $HCE_{\gamma} \downarrow$ \\
    \midrule
    Add  &0.802 & 0.742 & 0.066 & 0.827 & 0.880 & 1109      \\
    Concat  &0.813 & 0.751 & 0.063 & 0.832 & 0.886 & 1104\\
    \rowcolor[RGB]{235,235,235}
    TSA  & \textbf{0.823} & \textbf{0.763} & \textbf{0.059} & \textbf{0.838} & \textbf{0.892} & \textbf{1097} \\
  \bottomrule
\end{tabular}
\end{table}

\subsubsection{Dual- \emph{vs.} Single-size Input}
% As shown in the previous work \cite{qin2022highly}, using the dual-size input is essential for handling high-resolution images. In our analysis, we compare the performance of our model using single-scale inputs, including low-resolution (size: 256) and high-resolution (size: 1024) inputs, with our model using the dual-size input, as presented in Table 3.
% Indeed, it is intuitive that the HCE metric shows a significant improvement with high-resolution inputs as they retain more structural details. However, the maximal F-measure ($F_\beta^\text{max}\uparrow$) \cite{achanta2009frequency} and Mean Absolute Error (MAE, M \downarrow$) \cite{perazzi2012saliency} metrics do not show significant improvements, likely due to the limited receptive field of the model.
% By adopting the approach of using the dual-size input and sharing the backbone, we can retain the advantages of both scales while achieving comprehensive improvements in all metrics, with only a small increase in computational cost while still maintaining real-time inference speed.
%使用双尺度输入对于处理高分辨率图像是非常有必要的，因为它可以提供跨越更大范围的多尺度特征。我们在表3中比较了在UDUN中使用单尺度输入和双尺度输入时模型的性能。对比前两行我们发现，在高分辨率输入的情况下HCE得分有明显的改善，这归功于大分辨率包含更多的细节信息。通过双尺度输入，模型的表现在所有性能指标上都有明显的提升。尽管牺牲了一些推理速度，但是在共享主干的设计下使得模型只增加了少量的GFLOPs同时仍然保持实时的推理速度。
The dual-size input is essential for processing high-resolution segmentation because it enables the encoder to extract multiscale features over a wider range.
\tabref{tab:input} compares the performance of~\ourmodel~when using the single-size input and dual-size input. The results in the first two rows illustrate a significant improvement in HCE$_{\gamma}$ when using higher-resolution input, which is attributed to the greater amount of detailed information contained in the higher-resolution features. 
By utilizing the dual-size input, the performance of~\ourmodel~is further improved in terms of $F_\beta^\text{max}$, $M$, and HCE$_{\gamma}$. 
Despite sacrificing some inference speed, the shared backbone design enables our model to increase only a few GFLOPs while still keeping real-time inference.

\begin{figure}[t!]
\centering
\includegraphics[width=0.98\linewidth]{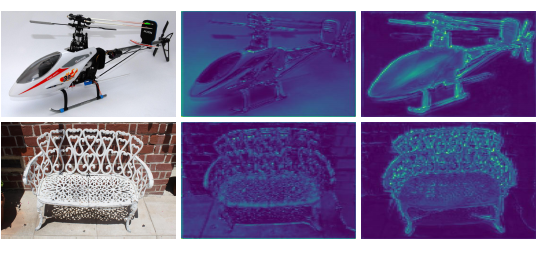}
  \put(-207,-5){{(a)}}
    \put(-125,-5){{(b)}}
    \put(-43,-5){{(c)}}
\caption{Comparison of feature maps before and after the filtering operation in the structure decoder. (a) Input image; (b) Before filtering; (c) After filtering.}\label{fig:filter}
\end{figure}

\begin{table}
\footnotesize
  \renewcommand{\arraystretch}{1}
  \renewcommand{\tabcolsep}{3.2mm}
  \caption{Comparisons of single- and dual-size inputs.}
  \label{tab:input}
  \begin{tabular}{c|cccccc}
    \toprule
    Input Sizes & $F_\beta^\text{max} \uparrow$ &  $M \downarrow$ & $HCE_{\gamma} \downarrow$ &
    GFLOPs & FPS \\
    \midrule
    256   &0.758 & 0.083 & 1486 & \textbf{16.4} & \textbf{80.2}     \\
    1024  &0.786 & 0.072 & 1138 & 131.9 &70.5    \\
    \rowcolor[RGB]{235,235,235}
    256\&1024  &\textbf{0.823} & \textbf{0.059} & \textbf{1097} & 142.3 &45.5  \\
  \bottomrule
\end{tabular}
\end{table}

%\noindent\textbf{Effectiveness of Decoupled Interior \$ Exterior label Supervision.}
%考虑放补充材料

\subsubsection{Large-scale Shallow Features}
%如何引出这个动机？IS-Net并没有1024，而且去掉1024HCE比不过IS-Net
% 
% 对于DIS任务而言，一些backbone 如resnet50对捕获很精细结构是不足的 due to continuous downsampling ，为了弥补这个缺陷
% 为了保留大尺度的结构完整性，所以我们设计了一个独立的large-scale（size:1024）路径，不需要经过heavy backbone仅仅几个卷积，可以以低代价获取高精度的细节。性能表现如表4所示，由于可以承载了更多结构信息，各个指标都显著提升。
% Based on our observations, we have found that large-scale features are essential for achieving ultra-high accuracy in Dichotomous Image Segmentation (DIS) tasks, as they enable retention of fine structures. Conversely, low-scale features are unable to effectively capture high-precision details. 
% To preserve the integrity of large-scale structures, we have designed a dedicated large-scale pathway with a size of 1024. This pathway does not require a heavy backbone and can achieve high-precision details through only a few convolutions, at a low computational cost. The performance results, as shown in Table \ref{tab: Large-scale Features}, demonstrate significant improvements across various metrics, due to the increased capacity to capture more structural information.
% 对于High-accuracy DIS任务而言，尽可能的保留大尺度的特征来捕捉低级的空间和纹理线索是非常有必要的。为了保留最大分辨率的特征并维持模型的低成本，我们嵌入了一个独立通道来直接从1024尺度的输入图像提取原尺度的浅层特征以输入到Structure解码器中而不经过backbone。如表6所示，该设计显著的提升了模型在所有指标上的表现，特征是在HCE指标上。这证明了大尺度低级别的特征包含的丰富的空间和纹理信息其可以帮助识别高分辨率目标中的精细结构。
For achieving highly accuracy masks, it is crucial to extremely preserve large-scale features that contain a wealth of low-level spatial and texture information. 
To ensure that~\ourmodel~retains the highest resolution features for feeding into the structure decoder while maintaining the lower computational cost, we embedded a stand-alone path that extracts the shallow features directly from the input image with a scale of 1024 without passing through the backbone (see \figref{fig:model}).
As shown in \tabref{tab:shallow_feat}, this design significantly boosts the performance of our model on all metrics, especially on the HCE$_{\gamma}$. 
It demonstrates the rich spatial and textural information available from large-scale low-level features that can facilitate the discrimination of fine-grained structures in high-resolution objects.

% 在structure decoder中，我们使用了purify operation 利用高级的trunk特征，它往往更聚焦前景的内部区域，除低级特征中的内部噪声和背景噪声，让网络更加关注前景结构信息.table6中展示了purify operation的性能，维持HCE指标的同时，增强了trunk的结构完整度从前5个指标可以显著看出。
% 同时，我们可视化了purify operation前后的特征，见图5，从图中可知，purify operation 之前网络只是关注整体轮廓，关注点在一些梯度大的边缘区域并夹杂者背景噪声，而使用了purify operation 之后，网络更加关注前景目标的structure包括前景的内部结构。
\subsubsection{Effectiveness of Filtering Operation}
% In the structure decoder, we have employed a filtering operation that utilizes high-level trunk features, which tend to focus more on the internal regions of the foreground. This operation removes internal noise and background noise from the low-level features, allowing the network to better focus on foreground structural information. The performance of the filtering operation is shown in \tabref{tab:purification}, where it is evident that while maintaining the HCE metric, the integrity of the trunk structure is significantly enhanced, as evident from the first 5 metrics.
% We also visualized the features before and after applying the filtering operation, as shown in  \figref{fig:purify}. From the figure, it can be observed that before the filtering operation, the network only focuses on the overall contours, with attention to areas with high gradients and background noise. However, after applying the filtering operation, the network pays more attention to the structure of the foreground targets, including the internal structure of the foreground.
The structure decoder embraces a simple feature filtering operation that leverages the intermediate trunk features in the trunk decoder to suppress background noise in lower-level features and focus on dominant information (refer to \eqnref{equ:filter}). 
\tabref{tab:filter} demonstrates the benefit of this operation.
It significantly enhances the holistic perception effect on the foreground area while keeping a superior HCE score. 
We also compare the visual features before and after adding the filtering operation. As shown in \figref{fig:filter}, after embedding the filtering operation, the feature maps are focused on the internal structure of the targets and effectively suppress substantial undesired background noise.

\begin{table}
\footnotesize
  \renewcommand{\arraystretch}{1}
  \renewcommand{\tabcolsep}{1.8mm}
  \caption{Effectiveness of large-scale shallow features for the structure decoder.}
  \label{tab:shallow_feat}
  \begin{tabular}{c|cccccc}
    \toprule
    Large-scale Features & $F_\beta^\text{max} \uparrow$ &  $F_\beta^\omega \uparrow$ & $M \downarrow$ &
    $S_\alpha \uparrow$ & $E_{\phi}^\text{m} \uparrow$ & $HCE_{\gamma} \downarrow$ \\
    \midrule
       &0.802 & 0.743 & 0.066 & 0.828 & 0.878 & 1157 \\
    \rowcolor[RGB]{235,235,235}
     \checkmark  &\textbf{0.823} & \textbf{0.763} & \textbf{0.059} & \textbf{0.838} & \textbf{0.892} & \textbf{1097} \\
  \bottomrule
\end{tabular}
\end{table}

\begin{table}
\footnotesize
  \renewcommand{\arraystretch}{1}
  \renewcommand{\tabcolsep}{1.9mm}
  \caption{Effectiveness of the filtering operation for the input features of the structure decoder.}
  \label{tab:filter}
  \begin{tabular}{c|cccccc}
    \toprule
    Filtering Operation & $F_\beta^\text{max} \uparrow$ &  $F_\beta^\omega \uparrow$ & $M \downarrow$ &
    $S_\alpha \uparrow$ & $E_{\phi}^\text{m} \uparrow$ & $HCE_{\gamma} \downarrow$ \\
    \midrule
       &0.813 & 0.752 & 0.062 & 0.835 & 0.887 & 1108\\
    \rowcolor[RGB]{235,235,235}
    \checkmark   &\textbf{0.823} & \textbf{0.763} & \textbf{0.059} & \textbf{0.838} & \textbf{0.892} & \textbf{1097} \\
  \bottomrule
\end{tabular}
\end{table}

\subsubsection{Performance of Different Backbones}
% 为了探究Shared and  Non-shared Backbone对性能的影响，我们将双尺度书入分别输入到两个同样的R50中得到两组分辨率的特征，如(Fig. 1(c))，性能展示在table7中，使用非共享方式达到了一个比较好的HCE指标，但是Fmax M指标却有所下降。这可能是因为输入高分辨率的backbone占据网络优化的主导，但是其感受野不足而导致目标定位不精确。采用共享的方式，可以使得一组backbone参数同时接受两种尺度的输入like zoom in and out 在定位目标trunk的同时不丢失精细的结构，同时减少了近乎一半的参数量使得网络更轻量。
% Shared and Non-shared Backbone are two different ways to use two scales of features in the network. In the Non-shared Backbone approach, each scale of feature is input into a separate ResNet50 backbone, and in the Shared Backbone approach, both scales of features are input into the same ResNet50 backbone.
% According to the results shown in Table 8, using the Non-shared Backbone approach achieved a good HCE score, but the Fmax M score decreased. This may be because the high-resolution backbone dominates the network optimization, but its receptive field is insufficient, leading to imprecise target localization. In contrast, using the Shared Backbone approach allows one set of backbone parameters to simultaneously process both scales of features, similar to zooming in and out, thus preserving fine-grained structures while localizing the trunk. Moreover, using the Shared Backbone approach reduces the number of parameters by almost half, making the network more lightweight.
\ourmodel~adopts a shared backbone to minimize the impact of model parameters from the dual-size input. 
To investigate the effect of the performance with shared and non-shared backbones, we compare the segmentation results on overall DIS-TE in the third and fourth rows of \tabref{tab:backbone}. 
Although the non-shared backbone style achieves a better HCE score, it has a decreasing trend in terms of $F_\beta^\text{max}$ and $M$ performance. 
In contrast, the shared backbone reduces the number of parameters by almost 50\%, resulting in a lightweight framework.
Moreover, we observe that the number of backbone parameters accounts for a large proportion of our model. 
As a result, we evaluate the results of our model with lighter backbones, including ResNet-18~\cite{he2016deep} and ResNet-34~\cite{he2016deep}. 
\tabref{tab:backbone} illustrates that when using ResNet-18, UDUN reaches a real-time inference speed of 65.3 \emph{fps} with only 12.33M parameters, while still outperforming IS-Net~\cite{qin2022highly} in all metrics on overall DIS-TE. 
The experimental results highlight the potential of our method for real-world applications.

% 消融实验是在验证集
%要不要展示R18 R34 R59在overall TE1-4六个指标都超过IS-net?  放补充材料？

\begin{table}
\footnotesize
  \renewcommand{\arraystretch}{1}
  \renewcommand{\tabcolsep}{2.05mm}
  \caption{Performance of~\ourmodel\ with diverse backbones on the overall DIS-TE test set.}
  \label{tab:backbone}
  \begin{tabular}{c|c|ccccc}
    \toprule
    Backbones &  Shared  & $F_\beta^\text{max} \uparrow$ &  $M \downarrow$ &
    $HCE_{\gamma} \downarrow$ & Params & FPS \\
    \midrule
          ResNet-18~\cite{he2016deep}  & \checkmark   &0.807  & 0.065 & 1009 & \textbf{12.33}M  &\textbf{65.3}  \\
          ResNet-34~\cite{he2016deep} & \checkmark   &0.816  & 0.061 & 995 & 22.45M  &55.0  \\
      
         % Res-50~\cite{he2016deep} (Non-shared) &0.810  & 0.063 & \textbf{1072} & 48.65  &42.3  \\
         ResNet-50~\cite{he2016deep} &    &0.826  &0.058  & \textbf{964} & 48.65M  &42.3  \\
         
    \rowcolor[RGB]{235,235,235}
    ResNet-50~\cite{he2016deep} &  \checkmark   &\textbf{0.831} & \textbf{0.057} & 977 & 25.05M &  45.5    \\
  \bottomrule
\end{tabular}
\end{table}

% \begin{table*}
%   \caption{Some Typical Commands}
%   \label{tab:commands}
%   \begin{tabular}{ccl}
%     \toprule
%     Command &A Number & Comments\\
%     \midrule
%     \texttt{{\char'134}author} & 100& Author \\
%     \texttt{{\char'134}table}& 300 & For tables\\
%     \texttt{{\char'134}table*}& 400& For wider tables\\
%     \bottomrule
%   \end{tabular}
% \end{table*}

% \begin{math}
%   \lim_{n\rightarrow \infty}x=0
% \end{math},

% \begin{equation}
%   \lim_{n\rightarrow \infty}x=0
% \end{equation}

% \begin{figure}[h]
%   \centering
%   \includegraphics[width=\linewidth]{sample-franklin}
%   \caption{1907 Franklin Model D roadster. Photograph by Harris \&
%     Ewing, Inc. [Public domain], via Wikimedia
%     Commons. (\url{https://goo.gl/VLCRBB}).}
%   \Description{A woman and a girl in white dresses sit in an open car.}
% \end{figure}

% \section{Acknowledgments}

% Identification of funding sources and other support, and thanks to individuals and groups that assisted in the research and the
% preparation of the work should be included in an acknowledgment
% section, which is placed just before the reference section in your document.

\section{Conclusion}

The main purpose of this paper is to explore how to segment highly accurate objects with intricate outlines and cumbersome structures in the DIS task. 
Unlike high-resolution task-specific binary image segmentation, the difficulty for DIS lies in simultaneously segmenting dominant regions and detailed structures with high accuracy. 
To address this challenge, we contribute a one-stage unite-divide-unite network (\ourmodel) that adopts a unite encoder to extract two sets of features at different scales from a shared backbone, then a divide-and-conquer strategy to refine and fuse the trunk and structure information in our trunk and structure decoders, respectively. 
Finally, a union decoder is introduced to integrate these two features for efficient segmentation of high-accuracy objects. 
Qualitative and quantitative experimental results demonstrate the effectiveness
and robustness of the proposed~\ourmodel.
We hope our model can be widely applied to various real-life scenarios, such as medical image analysis, image matting, and the art field. 


% 加了一个blance  参考文献排版就对齐了，但是警告还是在
\balance

\bibliographystyle{ACM-Reference-Format}
\bibliography{reference}


\begin{thebibliography}{46}

%%% the Web address.

\ifx \showCODEN    \undefined \def \showCODEN     #1{\unskip}     \fi
\ifx \showDOI      \undefined \def \showDOI       #1{#1}\fi
\ifx \showISBNx    \undefined \def \showISBNx     #1{\unskip}     \fi
\ifx \showISBNxiii \undefined \def \showISBNxiii  #1{\unskip}     \fi
\ifx \showISSN     \undefined \def \showISSN      #1{\unskip}     \fi
\ifx \showLCCN     \undefined \def \showLCCN      #1{\unskip}     \fi
\ifx \shownote     \undefined \def \shownote      #1{#1}          \fi
\ifx \showarticletitle \undefined \def \showarticletitle #1{#1}   \fi
\ifx \showURL      \undefined \def \showURL       {\relax}        \fi
% invisible to TeX
\providecommand\bibfield[2]{#2}
\providecommand\bibinfo[2]{#2}
\providecommand\natexlab[1]{#1}
\providecommand\showeprint[2][]{arXiv:#2}

\bibitem[Achanta et~al\mbox{.}(2009)]%
        {achanta2009frequency}
\bibfield{author}{\bibinfo{person}{Radhakrishna Achanta},
  \bibinfo{person}{Sheila Hemami}, \bibinfo{person}{Francisco Estrada}, {and}
  \bibinfo{person}{Sabine Susstrunk}.} \bibinfo{year}{2009}\natexlab{}.
\newblock \showarticletitle{Frequency-tuned salient region detection}. In
  \bibinfo{booktitle}{\emph{CVPR}}. \bibinfo{publisher}{IEEE},
  \bibinfo{address}{Miami, Florida, USA}, \bibinfo{pages}{1597--1604}.
\newblock


\bibitem[Chaurasia and Culurciello(2017)]%
        {chaurasia2017linknet}
\bibfield{author}{\bibinfo{person}{Abhishek Chaurasia} {and}
  \bibinfo{person}{Eugenio Culurciello}.} \bibinfo{year}{2017}\natexlab{}.
\newblock \showarticletitle{Linknet: Exploiting encoder representations for
  efficient semantic segmentation}. In \bibinfo{booktitle}{\emph{VCIP}}.
  \bibinfo{publisher}{IEEE}, \bibinfo{address}{Petersburg, FL, USA},
  \bibinfo{pages}{1--4}.
\newblock


\bibitem[Chen et~al\mbox{.}(2019)]%
        {chen2019collaborative}
\bibfield{author}{\bibinfo{person}{Wuyang Chen}, \bibinfo{person}{Ziyu Jiang},
  \bibinfo{person}{Zhangyang Wang}, \bibinfo{person}{Kexin Cui}, {and}
  \bibinfo{person}{Xiaoning Qian}.} \bibinfo{year}{2019}\natexlab{}.
\newblock \showarticletitle{Collaborative global-local networks for
  memory-efficient segmentation of ultra-high resolution images}. In
  \bibinfo{booktitle}{\emph{CVPR}}. \bibinfo{publisher}{IEEE},
  \bibinfo{address}{Long Beach, CA, USA}, \bibinfo{pages}{8924--8933}.
\newblock


\bibitem[Chen et~al\mbox{.}(2020)]%
        {chen2020global}
\bibfield{author}{\bibinfo{person}{Zuyao Chen}, \bibinfo{person}{Qianqian Xu},
  \bibinfo{person}{Runmin Cong}, {and} \bibinfo{person}{Qingming Huang}.}
  \bibinfo{year}{2020}\natexlab{}.
\newblock \showarticletitle{Global context-aware progressive aggregation
  network for salient object detection}. In \bibinfo{booktitle}{\emph{AAAI}}.
  \bibinfo{publisher}{AAAI Press}, \bibinfo{address}{New York, NY, USA},
  \bibinfo{pages}{10599--10606}.
\newblock


\bibitem[Cheng et~al\mbox{.}(2020)]%
        {cheng2020cascadepsp}
\bibfield{author}{\bibinfo{person}{Ho~Kei Cheng}, \bibinfo{person}{Jihoon
  Chung}, \bibinfo{person}{Yu-Wing Tai}, {and} \bibinfo{person}{Chi-Keung
  Tang}.} \bibinfo{year}{2020}\natexlab{}.
\newblock \showarticletitle{Cascadepsp: Toward class-agnostic and very
  high-resolution segmentation via global and local refinement}. In
  \bibinfo{booktitle}{\emph{CVPR}}. \bibinfo{publisher}{IEEE},
  \bibinfo{address}{Seattle, WA, USA}, \bibinfo{pages}{8890--8899}.
\newblock


\bibitem[Deng et~al\mbox{.}(2009)]%
        {deng2009imagenet}
\bibfield{author}{\bibinfo{person}{Jia Deng}, \bibinfo{person}{Wei Dong},
  \bibinfo{person}{Richard Socher}, \bibinfo{person}{Li-Jia Li},
  \bibinfo{person}{Kai Li}, {and} \bibinfo{person}{Li Fei-Fei}.}
  \bibinfo{year}{2009}\natexlab{}.
\newblock \showarticletitle{Imagenet: A large-scale hierarchical image
  database}. In \bibinfo{booktitle}{\emph{CVPR}}. \bibinfo{publisher}{IEEE},
  \bibinfo{address}{Miami, Florida, USA}, \bibinfo{pages}{248--255}.
\newblock


\bibitem[Fan et~al\mbox{.}(2017)]%
        {fan2017structure}
\bibfield{author}{\bibinfo{person}{Deng-Ping Fan}, \bibinfo{person}{Ming-Ming
  Cheng}, \bibinfo{person}{Yun Liu}, \bibinfo{person}{Tao Li}, {and}
  \bibinfo{person}{Ali Borji}.} \bibinfo{year}{2017}\natexlab{}.
\newblock \showarticletitle{Structure-measure: A new way to evaluate foreground
  maps}. In \bibinfo{booktitle}{\emph{ICCV}}. \bibinfo{publisher}{IEEE},
  \bibinfo{address}{Venice, Italy}, \bibinfo{pages}{4548--4557}.
\newblock


\bibitem[Fan et~al\mbox{.}(2018)]%
        {fan2018enhanced}
\bibfield{author}{\bibinfo{person}{Deng-Ping Fan}, \bibinfo{person}{Cheng
  Gong}, \bibinfo{person}{Yang Cao}, \bibinfo{person}{Bo Ren},
  \bibinfo{person}{Ming-Ming Cheng}, {and} \bibinfo{person}{Ali Borji}.}
  \bibinfo{year}{2018}\natexlab{}.
\newblock \showarticletitle{Enhanced-alignment measure for binary foreground
  map evaluation}. In \bibinfo{booktitle}{\emph{IJCAI}}.
  \bibinfo{publisher}{AAAI Press}, \bibinfo{address}{Stockholm, Sweden},
  \bibinfo{pages}{1--10}.
\newblock


\bibitem[Fan et~al\mbox{.}(2021a)]%
        {fan2021concealed}
\bibfield{author}{\bibinfo{person}{Deng-Ping Fan}, \bibinfo{person}{Ge-Peng
  Ji}, \bibinfo{person}{Ming-Ming Cheng}, {and} \bibinfo{person}{Ling Shao}.}
  \bibinfo{year}{2021}\natexlab{a}.
\newblock \showarticletitle{Concealed object detection}.
\newblock \bibinfo{journal}{\emph{IEEE TPAMI}} \bibinfo{volume}{44},
  \bibinfo{number}{10} (\bibinfo{year}{2021}), \bibinfo{pages}{6024--6042}.
\newblock


\bibitem[Fan et~al\mbox{.}(2022)]%
        {fan2022salient}
\bibfield{author}{\bibinfo{person}{Deng-Ping Fan}, \bibinfo{person}{Jing
  Zhang}, \bibinfo{person}{Gang Xu}, \bibinfo{person}{Ming-Ming Cheng}, {and}
  \bibinfo{person}{Ling Shao}.} \bibinfo{year}{2022}\natexlab{}.
\newblock \showarticletitle{Salient objects in clutter}.
\newblock \bibinfo{journal}{\emph{IEEE TPAMI}} \bibinfo{volume}{45},
  \bibinfo{number}{2} (\bibinfo{year}{2022}), \bibinfo{pages}{2344--2366}.
\newblock


\bibitem[Fan et~al\mbox{.}(2021b)]%
        {fan2021rethinking}
\bibfield{author}{\bibinfo{person}{Mingyuan Fan}, \bibinfo{person}{Shenqi Lai},
  \bibinfo{person}{Junshi Huang}, \bibinfo{person}{Xiaoming Wei},
  \bibinfo{person}{Zhenhua Chai}, \bibinfo{person}{Junfeng Luo}, {and}
  \bibinfo{person}{Xiaolin Wei}.} \bibinfo{year}{2021}\natexlab{b}.
\newblock \showarticletitle{Rethinking bisenet for real-time semantic
  segmentation}. In \bibinfo{booktitle}{\emph{CVPR}}.
  \bibinfo{publisher}{IEEE}, \bibinfo{address}{virtual},
  \bibinfo{pages}{9716--9725}.
\newblock


\bibitem[Gao et~al\mbox{.}(2019)]%
        {gao2019res2net}
\bibfield{author}{\bibinfo{person}{Shang-Hua Gao}, \bibinfo{person}{Ming-Ming
  Cheng}, \bibinfo{person}{Kai Zhao}, \bibinfo{person}{Xin-Yu Zhang},
  \bibinfo{person}{Ming-Hsuan Yang}, {and} \bibinfo{person}{Philip Torr}.}
  \bibinfo{year}{2019}\natexlab{}.
\newblock \showarticletitle{Res2net: A new multi-scale backbone architecture}.
\newblock \bibinfo{journal}{\emph{IEEE TPAMI}} \bibinfo{volume}{43},
  \bibinfo{number}{2} (\bibinfo{year}{2019}), \bibinfo{pages}{652--662}.
\newblock


\bibitem[Guo et~al\mbox{.}(2022)]%
        {guo2022isdnet}
\bibfield{author}{\bibinfo{person}{Shaohua Guo}, \bibinfo{person}{Liang Liu},
  \bibinfo{person}{Zhenye Gan}, \bibinfo{person}{Yabiao Wang},
  \bibinfo{person}{Wuhao Zhang}, \bibinfo{person}{Chengjie Wang},
  \bibinfo{person}{Guannan Jiang}, \bibinfo{person}{Wei Zhang},
  \bibinfo{person}{Ran Yi}, \bibinfo{person}{Lizhuang Ma}, {et~al\mbox{.}}}
  \bibinfo{year}{2022}\natexlab{}.
\newblock \showarticletitle{Isdnet: Integrating shallow and deep networks for
  efficient ultra-high resolution segmentation}. In
  \bibinfo{booktitle}{\emph{CVPR}}. \bibinfo{publisher}{IEEE},
  \bibinfo{address}{New Orleans, LA, USA}, \bibinfo{pages}{4361--4370}.
\newblock


\bibitem[He et~al\mbox{.}(2016)]%
        {he2016deep}
\bibfield{author}{\bibinfo{person}{Kaiming He}, \bibinfo{person}{Xiangyu
  Zhang}, \bibinfo{person}{Shaoqing Ren}, {and} \bibinfo{person}{Jian Sun}.}
  \bibinfo{year}{2016}\natexlab{}.
\newblock \showarticletitle{Deep residual learning for image recognition}. In
  \bibinfo{booktitle}{\emph{CVPR}}. \bibinfo{publisher}{IEEE},
  \bibinfo{address}{Las Vegas, NV, USA}, \bibinfo{pages}{770--778}.
\newblock


\bibitem[Hu et~al\mbox{.}(2022)]%
        {hu2022learning}
\bibfield{author}{\bibinfo{person}{Hanzhe Hu}, \bibinfo{person}{Yinbo Chen},
  \bibinfo{person}{Jiarui Xu}, \bibinfo{person}{Shubhankar Borse},
  \bibinfo{person}{Hong Cai}, \bibinfo{person}{Fatih Porikli}, {and}
  \bibinfo{person}{Xiaolong Wang}.} \bibinfo{year}{2022}\natexlab{}.
\newblock \showarticletitle{Learning implicit feature alignment function for
  semantic segmentation}. In \bibinfo{booktitle}{\emph{ECCV}}.
  \bibinfo{publisher}{Elsevier}, \bibinfo{address}{Tel Aviv, Israel},
  \bibinfo{pages}{487--505}.
\newblock


\bibitem[Li et~al\mbox{.}(2021)]%
        {li2021contexts}
\bibfield{author}{\bibinfo{person}{Qi Li}, \bibinfo{person}{Weixiang Yang},
  \bibinfo{person}{Wenxi Liu}, \bibinfo{person}{Yuanlong Yu}, {and}
  \bibinfo{person}{Shengfeng He}.} \bibinfo{year}{2021}\natexlab{}.
\newblock \showarticletitle{From contexts to locality: Ultra-high resolution
  image segmentation via locality-aware contextual correlation}. In
  \bibinfo{booktitle}{\emph{ICCV}}. \bibinfo{publisher}{IEEE},
  \bibinfo{address}{Montreal, QC, Canada}, \bibinfo{pages}{7252--7261}.
\newblock


\bibitem[Lin et~al\mbox{.}(2021)]%
        {lin2021real}
\bibfield{author}{\bibinfo{person}{Shanchuan Lin}, \bibinfo{person}{Andrey
  Ryabtsev}, \bibinfo{person}{Soumyadip Sengupta}, \bibinfo{person}{Brian~L
  Curless}, \bibinfo{person}{Steven~M Seitz}, {and} \bibinfo{person}{Ira
  Kemelmacher-Shlizerman}.} \bibinfo{year}{2021}\natexlab{}.
\newblock \showarticletitle{Real-time high-resolution background matting}. In
  \bibinfo{booktitle}{\emph{CVPR}}. \bibinfo{publisher}{IEEE},
  \bibinfo{address}{virtual}, \bibinfo{pages}{8762--8771}.
\newblock


\bibitem[Margolin et~al\mbox{.}(2014)]%
        {margolin2014evaluate}
\bibfield{author}{\bibinfo{person}{Ran Margolin}, \bibinfo{person}{Lihi
  Zelnik-Manor}, {and} \bibinfo{person}{Ayellet Tal}.}
  \bibinfo{year}{2014}\natexlab{}.
\newblock \showarticletitle{How to evaluate foreground maps?}. In
  \bibinfo{booktitle}{\emph{CVPR}}. \bibinfo{publisher}{IEEE},
  \bibinfo{address}{Columbus, OH, USA}, \bibinfo{pages}{248--255}.
\newblock


\bibitem[Mei et~al\mbox{.}(2021)]%
        {mei2021camouflaged}
\bibfield{author}{\bibinfo{person}{Haiyang Mei}, \bibinfo{person}{Ge-Peng Ji},
  \bibinfo{person}{Ziqi Wei}, \bibinfo{person}{Xin Yang},
  \bibinfo{person}{Xiaopeng Wei}, {and} \bibinfo{person}{Deng-Ping Fan}.}
  \bibinfo{year}{2021}\natexlab{}.
\newblock \showarticletitle{Camouflaged object segmentation with distraction
  mining}. In \bibinfo{booktitle}{\emph{CVPR}}. \bibinfo{publisher}{IEEE},
  \bibinfo{address}{virtual}, \bibinfo{pages}{8772--8781}.
\newblock


\bibitem[Pang et~al\mbox{.}(2022)]%
        {pang2022zoom}
\bibfield{author}{\bibinfo{person}{Youwei Pang}, \bibinfo{person}{Xiaoqi Zhao},
  \bibinfo{person}{Tian-Zhu Xiang}, \bibinfo{person}{Lihe Zhang}, {and}
  \bibinfo{person}{Huchuan Lu}.} \bibinfo{year}{2022}\natexlab{}.
\newblock \showarticletitle{Zoom in and out: A mixed-scale triplet network for
  camouflaged object detection}. In \bibinfo{booktitle}{\emph{CVPR}}.
  \bibinfo{publisher}{IEEE}, \bibinfo{address}{New Orleans, LA, USA},
  \bibinfo{pages}{2160--2170}.
\newblock


\bibitem[Perazzi et~al\mbox{.}(2012)]%
        {perazzi2012saliency}
\bibfield{author}{\bibinfo{person}{Federico Perazzi}, \bibinfo{person}{Philipp
  Kr{\"a}henb{\"u}hl}, \bibinfo{person}{Yael Pritch}, {and}
  \bibinfo{person}{Alexander Hornung}.} \bibinfo{year}{2012}\natexlab{}.
\newblock \showarticletitle{Saliency filters: Contrast based filtering for
  salient region detection}. In \bibinfo{booktitle}{\emph{CVPR}}.
  \bibinfo{publisher}{IEEE}, \bibinfo{address}{Providence, RI, USA},
  \bibinfo{pages}{733--740}.
\newblock


\bibitem[Pohlen et~al\mbox{.}(2017)]%
        {pohlen2017full}
\bibfield{author}{\bibinfo{person}{Tobias Pohlen}, \bibinfo{person}{Alexander
  Hermans}, \bibinfo{person}{Markus Mathias}, {and} \bibinfo{person}{Bastian
  Leibe}.} \bibinfo{year}{2017}\natexlab{}.
\newblock \showarticletitle{Full-resolution residual networks for semantic
  segmentation in street scenes}. In \bibinfo{booktitle}{\emph{CVPR}}.
  \bibinfo{publisher}{IEEE}, \bibinfo{address}{Honolulu, HI, USA},
  \bibinfo{pages}{4151--4160}.
\newblock


\bibitem[Qin et~al\mbox{.}(2022)]%
        {qin2022highly}
\bibfield{author}{\bibinfo{person}{Xuebin Qin}, \bibinfo{person}{Hang Dai},
  \bibinfo{person}{Xiaobin Hu}, \bibinfo{person}{Deng-Ping Fan},
  \bibinfo{person}{Ling Shao}, {and} \bibinfo{person}{Luc Van~Gool}.}
  \bibinfo{year}{2022}\natexlab{}.
\newblock \showarticletitle{Highly accurate dichotomous image segmentation}. In
  \bibinfo{booktitle}{\emph{ECCV}}. \bibinfo{publisher}{Springer},
  \bibinfo{address}{Tel Aviv, Israel}, \bibinfo{pages}{38--56}.
\newblock


\bibitem[Qin et~al\mbox{.}(2020)]%
        {qin2020u2}
\bibfield{author}{\bibinfo{person}{Xuebin Qin}, \bibinfo{person}{Zichen Zhang},
  \bibinfo{person}{Chenyang Huang}, \bibinfo{person}{Masood Dehghan},
  \bibinfo{person}{Osmar~R Zaiane}, {and} \bibinfo{person}{Martin Jagersand}.}
  \bibinfo{year}{2020}\natexlab{}.
\newblock \showarticletitle{U2-Net: Going deeper with nested U-structure for
  salient object detection}.
\newblock \bibinfo{journal}{\emph{Pattern recognition}}  \bibinfo{volume}{106}
  (\bibinfo{year}{2020}), \bibinfo{pages}{107404}.
\newblock


\bibitem[Qin et~al\mbox{.}(2019)]%
        {qin2019basnet}
\bibfield{author}{\bibinfo{person}{Xuebin Qin}, \bibinfo{person}{Zichen Zhang},
  \bibinfo{person}{Chenyang Huang}, \bibinfo{person}{Chao Gao},
  \bibinfo{person}{Masood Dehghan}, {and} \bibinfo{person}{Martin Jagersand}.}
  \bibinfo{year}{2019}\natexlab{}.
\newblock \showarticletitle{Basnet: Boundary-aware salient object detection}.
  In \bibinfo{booktitle}{\emph{CVPR}}. \bibinfo{publisher}{IEEE},
  \bibinfo{address}{Long Beach, CA, USA}, \bibinfo{pages}{7479--7489}.
\newblock


\bibitem[Romera et~al\mbox{.}(2017)]%
        {romera2017erfnet}
\bibfield{author}{\bibinfo{person}{Eduardo Romera}, \bibinfo{person}{Jose~M
  Alvarez}, \bibinfo{person}{Luis~M Bergasa}, {and} \bibinfo{person}{Roberto
  Arroyo}.} \bibinfo{year}{2017}\natexlab{}.
\newblock \showarticletitle{Erfnet: Efficient residual factorized convnet for
  real-time semantic segmentation}.
\newblock \bibinfo{journal}{\emph{IEEE TITS}} \bibinfo{volume}{19},
  \bibinfo{number}{1} (\bibinfo{year}{2017}), \bibinfo{pages}{263--272}.
\newblock


\bibitem[Ronneberger et~al\mbox{.}(2015)]%
        {ronneberger2015u}
\bibfield{author}{\bibinfo{person}{Olaf Ronneberger}, \bibinfo{person}{Philipp
  Fischer}, {and} \bibinfo{person}{Thomas Brox}.}
  \bibinfo{year}{2015}\natexlab{}.
\newblock \showarticletitle{U-net: Convolutional networks for biomedical image
  segmentation}. In \bibinfo{booktitle}{\emph{MICCAI}},
  Vol.~\bibinfo{volume}{9351}. \bibinfo{publisher}{Springer},
  \bibinfo{address}{Munich, Germany}, \bibinfo{pages}{234--241}.
\newblock


\bibitem[Shen et~al\mbox{.}(2022)]%
        {shen2022high}
\bibfield{author}{\bibinfo{person}{Tiancheng Shen}, \bibinfo{person}{Yuechen
  Zhang}, \bibinfo{person}{Lu Qi}, \bibinfo{person}{Jason Kuen},
  \bibinfo{person}{Xingyu Xie}, \bibinfo{person}{Jianlong Wu},
  \bibinfo{person}{Zhe Lin}, {and} \bibinfo{person}{Jiaya Jia}.}
  \bibinfo{year}{2022}\natexlab{}.
\newblock \showarticletitle{High quality segmentation for ultra high-resolution
  images}. In \bibinfo{booktitle}{\emph{CVPR}}. \bibinfo{publisher}{IEEE},
  \bibinfo{address}{New Orleans, LA, USA}, \bibinfo{pages}{1310--1319}.
\newblock


\bibitem[Tang et~al\mbox{.}(2021)]%
        {tang2021disentangled}
\bibfield{author}{\bibinfo{person}{Lv Tang}, \bibinfo{person}{Bo Li},
  \bibinfo{person}{Yijie Zhong}, \bibinfo{person}{Shouhong Ding}, {and}
  \bibinfo{person}{Mofei Song}.} \bibinfo{year}{2021}\natexlab{}.
\newblock \showarticletitle{Disentangled high quality salient object
  detection}. In \bibinfo{booktitle}{\emph{ICCV}}. \bibinfo{publisher}{IEEE},
  \bibinfo{address}{Montreal, QC, Canada}, \bibinfo{pages}{3580--3590}.
\newblock


\bibitem[Tian et~al\mbox{.}(2022)]%
        {tian2022kine}
\bibfield{author}{\bibinfo{person}{Yang Tian}, \bibinfo{person}{Hualong Bai},
  \bibinfo{person}{Shengdong Zhao}, \bibinfo{person}{Chi-Wing Fu},
  \bibinfo{person}{Chun Yu}, \bibinfo{person}{Haozhao Qin},
  \bibinfo{person}{Qiong Wang}, {and} \bibinfo{person}{Pheng-Ann Heng}.}
  \bibinfo{year}{2022}\natexlab{}.
\newblock \showarticletitle{Kine-Appendage: Enhancing Freehand VR Interaction
  Through Transformations of Virtual Appendages}.
\newblock \bibinfo{journal}{\emph{IEEE TVCG}} \bibinfo{volume}{1},
  \bibinfo{number}{1} (\bibinfo{year}{2022}), \bibinfo{pages}{1--17}.
\newblock


\bibitem[Valanarasu and Patel(2022)]%
        {valanarasu2022unext}
\bibfield{author}{\bibinfo{person}{Jeya Maria~Jose Valanarasu} {and}
  \bibinfo{person}{Vishal~M Patel}.} \bibinfo{year}{2022}\natexlab{}.
\newblock \showarticletitle{Unext: Mlp-based rapid medical image segmentation
  network}. In \bibinfo{booktitle}{\emph{MICCAI}}.
  \bibinfo{publisher}{Springer}, \bibinfo{address}{Singapore},
  \bibinfo{pages}{23--33}.
\newblock


\bibitem[Wang et~al\mbox{.}(2020)]%
        {wang2020deep}
\bibfield{author}{\bibinfo{person}{Jingdong Wang}, \bibinfo{person}{Ke Sun},
  \bibinfo{person}{Tianheng Cheng}, \bibinfo{person}{Borui Jiang},
  \bibinfo{person}{Chaorui Deng}, \bibinfo{person}{Yang Zhao},
  \bibinfo{person}{Dong Liu}, \bibinfo{person}{Yadong Mu},
  \bibinfo{person}{Mingkui Tan}, \bibinfo{person}{Xinggang Wang},
  {et~al\mbox{.}}} \bibinfo{year}{2020}\natexlab{}.
\newblock \showarticletitle{Deep high-resolution representation learning for
  visual recognition}.
\newblock \bibinfo{journal}{\emph{IEEE TPAMI}} \bibinfo{volume}{43},
  \bibinfo{number}{10} (\bibinfo{year}{2020}), \bibinfo{pages}{3349--3364}.
\newblock


\bibitem[Wang et~al\mbox{.}(2021)]%
        {wang2021salient}
\bibfield{author}{\bibinfo{person}{Wenguan Wang}, \bibinfo{person}{Qiuxia Lai},
  \bibinfo{person}{Huazhu Fu}, \bibinfo{person}{Jianbing Shen},
  \bibinfo{person}{Haibin Ling}, {and} \bibinfo{person}{Ruigang Yang}.}
  \bibinfo{year}{2021}\natexlab{}.
\newblock \showarticletitle{Salient object detection in the deep learning era:
  An in-depth survey}.
\newblock \bibinfo{journal}{\emph{IEEE TPAMI}} \bibinfo{volume}{44},
  \bibinfo{number}{6} (\bibinfo{year}{2021}), \bibinfo{pages}{3239--3259}.
\newblock


\bibitem[Wei et~al\mbox{.}(2020a)]%
        {wei2020f3net}
\bibfield{author}{\bibinfo{person}{Jun Wei}, \bibinfo{person}{Shuhui Wang},
  {and} \bibinfo{person}{Qingming Huang}.} \bibinfo{year}{2020}\natexlab{a}.
\newblock \showarticletitle{F$^3$Net: fusion, feedback and focus for salient
  object detection}. In \bibinfo{booktitle}{\emph{AAAI}}.
  \bibinfo{publisher}{AAAI Press}, \bibinfo{address}{New York, NY, USA},
  \bibinfo{pages}{12321--12328}.
\newblock


\bibitem[Wei et~al\mbox{.}(2020b)]%
        {wei2020label}
\bibfield{author}{\bibinfo{person}{Jun Wei}, \bibinfo{person}{Shuhui Wang},
  \bibinfo{person}{Zhe Wu}, \bibinfo{person}{Chi Su}, \bibinfo{person}{Qingming
  Huang}, {and} \bibinfo{person}{Qi Tian}.} \bibinfo{year}{2020}\natexlab{b}.
\newblock \showarticletitle{Label decoupling framework for salient object
  detection}. In \bibinfo{booktitle}{\emph{CVPR}}. \bibinfo{publisher}{IEEE},
  \bibinfo{address}{Seattle, WA, USA}, \bibinfo{pages}{13025--13034}.
\newblock


\bibitem[Xie et~al\mbox{.}(2022)]%
        {xie2022pyramid}
\bibfield{author}{\bibinfo{person}{Chenxi Xie}, \bibinfo{person}{Changqun Xia},
  \bibinfo{person}{Mingcan Ma}, \bibinfo{person}{Zhirui Zhao},
  \bibinfo{person}{Xiaowu Chen}, {and} \bibinfo{person}{Jia Li}.}
  \bibinfo{year}{2022}\natexlab{}.
\newblock \showarticletitle{Pyramid grafting network for one-stage high
  resolution saliency detection}. In \bibinfo{booktitle}{\emph{CVPR}}.
  \bibinfo{publisher}{IEEE}, \bibinfo{address}{New Orleans, LA, USA},
  \bibinfo{pages}{11717--11726}.
\newblock


\bibitem[Yu et~al\mbox{.}(2018)]%
        {yu2018bisenet}
\bibfield{author}{\bibinfo{person}{Changqian Yu}, \bibinfo{person}{Jingbo
  Wang}, \bibinfo{person}{Chao Peng}, \bibinfo{person}{Changxin Gao},
  \bibinfo{person}{Gang Yu}, {and} \bibinfo{person}{Nong Sang}.}
  \bibinfo{year}{2018}\natexlab{}.
\newblock \showarticletitle{Bisenet: Bilateral segmentation network for
  real-time semantic segmentation}. In \bibinfo{booktitle}{\emph{ECCV}}.
  \bibinfo{publisher}{Elsevier}, \bibinfo{address}{Munich, Germany},
  \bibinfo{pages}{325--341}.
\newblock


\bibitem[Yu et~al\mbox{.}(2021)]%
        {yu2021mask}
\bibfield{author}{\bibinfo{person}{Qihang Yu}, \bibinfo{person}{Jianming
  Zhang}, \bibinfo{person}{He Zhang}, \bibinfo{person}{Yilin Wang},
  \bibinfo{person}{Zhe Lin}, \bibinfo{person}{Ning Xu}, \bibinfo{person}{Yutong
  Bai}, {and} \bibinfo{person}{Alan Yuille}.} \bibinfo{year}{2021}\natexlab{}.
\newblock \showarticletitle{Mask guided matting via progressive refinement
  network}. In \bibinfo{booktitle}{\emph{CVPR}}. \bibinfo{publisher}{IEEE},
  \bibinfo{address}{virtual}, \bibinfo{pages}{1154--1163}.
\newblock


\bibitem[Zeng et~al\mbox{.}(2019)]%
        {zeng2019towards}
\bibfield{author}{\bibinfo{person}{Yi Zeng}, \bibinfo{person}{Pingping Zhang},
  \bibinfo{person}{Jianming Zhang}, \bibinfo{person}{Zhe Lin}, {and}
  \bibinfo{person}{Huchuan Lu}.} \bibinfo{year}{2019}\natexlab{}.
\newblock \showarticletitle{Towards high-resolution salient object detection}.
  In \bibinfo{booktitle}{\emph{ICCV}}. \bibinfo{publisher}{IEEE},
  \bibinfo{address}{Seoul, Korea (South)}, \bibinfo{pages}{7234--7243}.
\newblock


\bibitem[Zhang et~al\mbox{.}(2021a)]%
        {zhang2021auto}
\bibfield{author}{\bibinfo{person}{Miao Zhang}, \bibinfo{person}{Tingwei Liu},
  \bibinfo{person}{Yongri Piao}, \bibinfo{person}{Shunyu Yao}, {and}
  \bibinfo{person}{Huchuan Lu}.} \bibinfo{year}{2021}\natexlab{a}.
\newblock \showarticletitle{Auto-msfnet: Search multi-scale fusion network for
  salient object detection}. In \bibinfo{booktitle}{\emph{Multimedia}}.
  \bibinfo{publisher}{ACM}, \bibinfo{address}{Chengdu, China},
  \bibinfo{pages}{667--676}.
\newblock


\bibitem[Zhang et~al\mbox{.}(2021b)]%
        {zhang2021looking}
\bibfield{author}{\bibinfo{person}{Pingping Zhang}, \bibinfo{person}{Wei Liu},
  \bibinfo{person}{Yi Zeng}, \bibinfo{person}{Yinjie Lei}, {and}
  \bibinfo{person}{Huchuan Lu}.} \bibinfo{year}{2021}\natexlab{b}.
\newblock \showarticletitle{Looking for the detail and context devils:
  High-resolution salient object detection}.
\newblock \bibinfo{journal}{\emph{IEEE TIP}}  \bibinfo{volume}{30}
  (\bibinfo{year}{2021}), \bibinfo{pages}{3204--3216}.
\newblock


\bibitem[Zhao et~al\mbox{.}(2020)]%
        {zhao2020suppress}
\bibfield{author}{\bibinfo{person}{Xiaoqi Zhao}, \bibinfo{person}{Youwei Pang},
  \bibinfo{person}{Lihe Zhang}, \bibinfo{person}{Huchuan Lu}, {and}
  \bibinfo{person}{Lei Zhang}.} \bibinfo{year}{2020}\natexlab{}.
\newblock \showarticletitle{Suppress and balance: A simple gated network for
  salient object detection}. In \bibinfo{booktitle}{\emph{ECCV}}.
  \bibinfo{publisher}{Elsevier}, \bibinfo{address}{Glasgow, UK},
  \bibinfo{pages}{35--51}.
\newblock


\bibitem[Zhao et~al\mbox{.}(2022)]%
        {zhao2022trasetr}
\bibfield{author}{\bibinfo{person}{Zixu Zhao}, \bibinfo{person}{Yueming Jin},
  {and} \bibinfo{person}{Pheng-Ann Heng}.} \bibinfo{year}{2022}\natexlab{}.
\newblock \showarticletitle{Trasetr: track-to-segment transformer with
  contrastive query for instance-level instrument segmentation in robotic
  surgery}. In \bibinfo{booktitle}{\emph{ICRA}}. \bibinfo{publisher}{IEEE},
  \bibinfo{address}{Philadelphia, PA, USA}, \bibinfo{pages}{11186--11193}.
\newblock


\bibitem[Zhao et~al\mbox{.}(2021)]%
        {zhao2021complementary}
\bibfield{author}{\bibinfo{person}{Zhirui Zhao}, \bibinfo{person}{Changqun
  Xia}, \bibinfo{person}{Chenxi Xie}, {and} \bibinfo{person}{Jia Li}.}
  \bibinfo{year}{2021}\natexlab{}.
\newblock \showarticletitle{Complementary trilateral decoder for fast and
  accurate salient object detection}. In
  \bibinfo{booktitle}{\emph{Multimedia}}. \bibinfo{publisher}{ACM},
  \bibinfo{address}{Chengdu, China}, \bibinfo{pages}{4967--4975}.
\newblock


\bibitem[Zhu et~al\mbox{.}(2022)]%
        {zhu2022can}
\bibfield{author}{\bibinfo{person}{Hongwei Zhu}, \bibinfo{person}{Peng Li},
  \bibinfo{person}{Haoran Xie}, \bibinfo{person}{Xuefeng Yan},
  \bibinfo{person}{Dong Liang}, \bibinfo{person}{Dapeng Chen},
  \bibinfo{person}{Mingqiang Wei}, {and} \bibinfo{person}{Jing Qin}.}
  \bibinfo{year}{2022}\natexlab{}.
\newblock \showarticletitle{I can find you! Boundary-guided separated attention
  network for camouflaged object detection}. In
  \bibinfo{booktitle}{\emph{AAAI}}. \bibinfo{publisher}{AAAI Press},
  \bibinfo{address}{New York, NY, USA}, \bibinfo{pages}{3608--3616}.
\newblock


\bibitem[Zhuge et~al\mbox{.}(2023)]%
        {zhuge2022salient}
\bibfield{author}{\bibinfo{person}{Mingchen Zhuge}, \bibinfo{person}{Deng-Ping
  Fan}, \bibinfo{person}{Nian Liu}, \bibinfo{person}{Dingwen Zhang},
  \bibinfo{person}{Dong Xu}, {and} \bibinfo{person}{Ling Shao}.}
  \bibinfo{year}{2023}\natexlab{}.
\newblock \showarticletitle{Salient object detection via integrity learning}.
\newblock \bibinfo{journal}{\emph{IEEE TPAMI}} \bibinfo{volume}{45},
  \bibinfo{number}{3} (\bibinfo{year}{2023}), \bibinfo{pages}{3738--3752}.
\newblock


\end{thebibliography}

% \appendix


\end{document}